\documentclass{article}

\usepackage[final]{corl_2021} 
\usepackage{times}
\usepackage{bm}
\usepackage{amsmath} 
\usepackage{amssymb}  
\usepackage{amsthm} 
\usepackage{tabularx, colortbl}
\usepackage{xcolor} 
\usepackage{graphicx,amsmath}
\usepackage[ruled,vlined]{algorithm2e}
\usepackage[numbers]{natbib}
\usepackage{multicol}
\usepackage{tikz}
\usetikzlibrary{arrows.meta}
\usepackage{siunitx}
\usepackage{sidecap}
\usepackage[font=small,labelfont=small]{caption}
\usepackage{color}

\graphicspath{{Figures/}}

\newcommand{\trsp}{\mathsf{T}}

\DeclareMathOperator*{\argmin}{argmin} 

\definecolor{lightblue}{rgb}{0., 0.35, 0.6}
\definecolor{lightred}{rgb}{0.8, 0.22, 0.29}
\definecolor{lightyellow}{rgb}{1.0, 0.84, 0.0}
\definecolor{navyblue}{rgb}{0.0, 0.0, 0.5}
\definecolor{navypurple}{rgb}{0.4, 0.0, 0.61}
\definecolor{lightgray}{rgb}{0.7, 0.7, 0.7}
\definecolor{deepskyblue}{rgb}{0., 0.75, 1.0}
\definecolor{crimson}{rgb}{0.86, 0.08, 0.23}

\DeclareRobustCommand{\blueEucl}{\raisebox{2pt}{\tikz{\draw[lightblue,solid,line width = 1.1pt](0,0) -- (4mm,0);}}}
\DeclareRobustCommand{\redGeod}{\raisebox{2pt}{\tikz{\draw[lightred,solid,line width = 1.1pt](0,0) -- (4mm,0);}}}
\DeclareRobustCommand{\bluedemo}{\raisebox{2pt}{\tikz{\draw[black!50!blue,solid,line width = 1.1pt](0,0) -- (4mm,0);}}}
\DeclareRobustCommand{\redrepro}{\raisebox{2pt}{\tikz{\draw[red,solid,line width = 1.1pt](0,0) -- (4mm,0);}}}
\DeclareRobustCommand{\lightredellipse}{\tikz{ \filldraw[color=white, fill=red!40, thick] (1.5,0) ellipse (.35 and 0.1);}}
\DeclareRobustCommand{\purplerepro}{\raisebox{2pt}{\tikz{\draw[navypurple,solid,line width = 1.1pt](0,0) -- (4mm,0);}}}
\DeclareRobustCommand{\purpleellipse}{\tikz{ \filldraw[color=white, fill=navypurple, thick] (1.5,0) ellipse (.35 and 0.1);}}
\DeclareRobustCommand{\blackcircle}{\tikz{ \filldraw[color=white, fill=black!80, thick](0,0) circle (.075);}}
\DeclareRobustCommand{\yellowcircle}{\tikz{ \filldraw[color=white, fill=lightyellow, thick](0,0) circle (.075);}}
\DeclareRobustCommand{\bluecircle}{\tikz{ \filldraw[color=white, fill=navyblue, thick](0,0) circle (.075);}}
\DeclareRobustCommand{\blackrepro}{\raisebox{2pt}{\tikz{\draw[black,solid,line width = 1.1pt](0,0) -- (4mm,0);}}}
\DeclareRobustCommand{\graydemo}{\raisebox{2pt}{\tikz{\draw[lightgray,solid,line width = 1.1pt](0,0) -- (4mm,0);}}}
\DeclareRobustCommand{\skybluedemo}{\raisebox{2pt}{\tikz{\draw[deepskyblue,solid,line width = 1.1pt](0,0) -- (4mm,0);}}}
\DeclareRobustCommand{\redcircle}{\tikz{ \filldraw[color=white, fill=crimson, thick](0,0) circle (.075);}}

\makeatletter
\renewcommand{\@noticestring}{\textsuperscript{\ensuremath{*}}Equal contribution. \\\@conferenceordinal\/ Conference on Robot Learning (CoRL \@conferenceyear), \@conferencelocation.}
\makeatother

\title{Orientation Probabilistic Movement Primitives \\on Riemannian Manifolds}

\author{Leonel Rozo\textsuperscript{\ensuremath{*1}}
	\quad
	Vedant Dave\textsuperscript{\ensuremath{*1,2}}
	\\
	\textsuperscript{\ensuremath{1}}Bosch Center for Artificial Intelligence. Renningen, Germany.
	\\
	\textsuperscript{\ensuremath{2}}University of Leoben. Leoben, Austria.
	\\
	\href{mailto:leonel.rozo@de.bosch.com}{\textrm{leonel.rozo@de.bosch.com}} 
	\quad
	\href{mailto:vedant.dave@unileoben.ac.at}{\textrm{vedant.dave@unileoben.ac.at}} 
}

\begin{document}
	\maketitle
	
	
	\begin{abstract}
		Learning complex robot motions necessarily demands to have models that are able to encode and retrieve full-pose trajectories when tasks are defined in operational spaces. Probabilistic movement primitives (ProMPs) stand out as a principled approach that models trajectory distributions learned from demonstrations. ProMPs allow for trajectory modulation and blending to achieve better generalization to novel situations. However, when ProMPs are employed in operational space, their original formulation does not directly apply to full-pose movements including rotational trajectories described by quaternions. This paper proposes a Riemannian formulation of ProMPs that enables encoding and retrieving of quaternion trajectories. Our method builds on Riemannian manifold theory, and exploits multilinear geodesic regression for estimating the ProMPs parameters. This novel approach makes ProMPs a suitable model for learning complex full-pose robot motion patterns. Riemannian ProMPs are tested on toy examples to illustrate their workflow, and on real learning-from-demonstration experiments. 
	\end{abstract}
	
	\keywords{Learning from Demonstration, Riemannian Manifolds} 
	
	\section{Introduction}
	\label{sec:intro}
	For robots to perform autonomously in unstructured environments, they need to learn to interact with their surroundings.
	To do so, robots may rely on a library of skills to execute simple motions or perform complicated tasks as a composition of several skills~\cite{Schaal03}. 
	A well-established way to learn motion skills is via human examples, a.k.a. learning from demonstrations (LfD)~\cite{Billard16}. 
	This entails a human expert showing once or several times a specific motion to be imitated by a robot.
	In this context, several models to encode demonstrations and synthesize motions have been proposed: Dynamical-systems approaches~\cite{Ijspeert13dmp}, probabilistic methods~\cite{paraschos2018probabilistic,calinon2016tutorial}, and lately, neural networks~\cite{Yunus2019CNMPs,Bahl20:NDPs}. 
	
	Probabilistic movement primitives (ProMPs)~\cite{paraschos2018probabilistic}, Stable Estimators of Dynamical Systems (SEDS)~\cite{Khansari2011seds}, Task-Parameterized Gaussian Mixture Models (TP-GMMs)~\cite{calinon2016tutorial}, Kernelized MPs (KMPs)~\cite{Huang2019KMPs} and Conditional Neural MPs (CNMPs)~\cite{Yunus2019CNMPs} are some of the recent probabilistic models to represent motion primitives (MPs). 
	While some of them were originally proposed to learn joint space motions (e.g., ProMPs), others have mainly focused on MPs in task space (e.g., TP-GMM), assuming that certain tasks, such as manipulation, may be more easily represented in Cartesian coordinates. 
	The latter approach comes with an additional challenge, namely, encoding and synthesizing end-effector orientation trajectories.
	This problem is often overlooked, but the need of robots performing complicated tasks makes it imperative to extend MPs models to handle orientation data. 
	
	Orientation representation in robotics comes in different ways such as Euler angles, rotation matrices and quaternions. Euler angles are a minimal and intuitive representation, which however is not unique. They are also known to be undesirable in feedback control due to singularities~\cite{Yuan88quaternionCtrl,Hemingway2018:gimballock}. Rotation matrices are often impractical due to their number of parameters. In contrast, quaternions are a nearly-minimal representation and provide strong stability guarantees in close-loop orientation control~\cite{Yuan88quaternionCtrl}. Despite their antipodality (i.e., each rotation is represented by two antipodal points on the sphere $\mathcal{S}^3$), quaternions have gained interest in robot learning, control, and optimization due to their favorable properties~\cite{Yuan88quaternionCtrl,Pastor11dmpOri,Silverio15tpgmm,Jaquier19CoRLa}. \citet{Pastor11dmpOri} and \citet{Ude14ICRA}~pioneered works to learn quaternion MPs, where the classic DMP~\cite{Ijspeert13dmp} is formulated to encode quaternion trajectories for robot motion generation. Its stability guarantees were later improved in~\cite{Koutras19CoRL}, while~\citet{Saveriano19mergingDMPs} used this new formulation into a single DMP framework to learn full-pose end-effector trajectories. However, DMPs do not encode the demonstrations variability, which is often exploited to retrieve collision-free trajectories or to design robot controllers~\cite{paraschos2018probabilistic,Yunus2019CNMPs}.       
	
	\citet{Silverio15tpgmm} proposed the first extension of TP-GMM for learning quaternion trajectories, which relies on unit-norm approximations. This was resolved in~\cite{Zeestraten17riemannian}, where a Riemannian manifold formulation for both motion learning and reproduction was introduced. Despite TP-GMM offers good extrapolation capabilities due to its task-parameterized formulation, none of the foregoing works provides via-point modulation or trajectories blending. These issues were addressed by KMP~\cite{Huang2019KMPs} for position data, whose model builds on a previously-learned probabilistic MP, such as TP-GMM. KMP was then extended to handle orientation data~\cite{Huang20TRO}, via a projection onto an Euclidean space using a fixed quaternion reference, i.e. a single Euclidean tangent space. Thus, model learning, via-point adaptation, and skill reproduction take place in Euclidean space, ignoring the intrinsic quaternions geometry. Note that such approximations lead to data and model distortions~\cite{Zeestraten17riemannian}.  
	
	Although ProMPs have been used to learn Cartesian movements, their formulation does not handle quaternion trajectories. A possible solution would entail unit-norm approximations as in~\cite{Silverio15tpgmm}, but this approach fully ignores the geometry of the quaternions space and may lead to inaccurate models. An alternative and more sound solution relies on a Riemannian manifold formulation of ProMP, in the same spirit as~\cite{Zeestraten17riemannian}. However, two main difficulties arise: (\emph{i}) learning of the model parameters does not accept a closed-form solution, and (\emph{ii}) trajectory retrieval is constrained to lie on the sphere $\mathcal{S}^3$. We here provide solutions to these and related problems, which lead to the first ProMP framework that makes possible encoding and reproducing full-pose end-effector trajectories. In contrast to TP-GMM methods, our approach provides via-point modulation and blending capabilities, which are naturally inherited from the original ProMP. Unlike KMP, ProMP is a compact and standalone model, meaning that learning and reproduction do not rely on previously-learned MPs. Moreover, our approach is not prone to inaccuracies arising from geometry-unaware operations.      
	
	Specifically, we introduce a Riemannian manifold approach to learn orientation motion primitives using ProMP. 
	Our extension builds on the classic ProMP~\cite{paraschos2018probabilistic} (summarized in~\S~\ref{sec:background}), and considers the space of quaternions as a Riemannian manifold. 
	We propose to estimate the ProMP parameters using multivariate geodesic regression (see~\S~\ref{sec:quaternion_promps}), and we show how trajectory retrieval, modulation of the trajectory distributions, and MPs blending are all possible via a Riemannian probabilistic formulation. 
	In~\S~\ref{sec:experiments}, we illustrate our approach on the unit-sphere manifold $\mathcal{S}^2$, and we learn realistic motion skills on a $7$-DoF robotic manipulator featuring complex full-pose trajectories on $\mathbb{R}^3 \times \mathcal{S}^3$. 
	

	\section{Background}
	\label{sec:background}
	
	\subsection{ProMPs}
	\label{subsec:promps}
	Probabilistic Movement Primitives (ProMPs)~\cite{paraschos2018probabilistic} is a probabilistic framework for learning and synthesizing robot motion skills. ProMPs represent a trajectory distribution by a set of basis functions. 
	Its probabilistic formulation enables movement modulation, parallel movement activation, and exploitation of variance information in robot control.
	Formally, a single movement trajectory is denoted by $\bm{\tau} = \{\bm{y}_t\}_{t=1}^T$, where $\bm{y}_t$ is a $d$-dimensional vector representing either a joint configuration or a Cartesian position at time step $t$ (additional time derivatives of $\bm{y}$ may also be considered). 
	Each point of the trajectory $\bm{\tau}$ is represented as a linear basis function model
	\begin{equation} \label{eq:ClassicProMP}
		\bm{y}_t =  \bm{\Psi}_t \bm{w} + \bm{\epsilon}_y \Rightarrow \mathcal{P}(\bm{y}_t | \bm{w}) = \mathcal{N}(\bm{y}_t | \bm{\Psi}_t \bm{w}, \bm{\Sigma}_{\bm{y}}) ,
	\end{equation} 
	where $\bm{w}$ is a $dN_{\phi}$-dimensional weight vector, $\bm{\Psi}_t$ is a fixed $d \times dN_{\phi}$-dimensional block diagonal matrix containing $N_{\phi}$ time-dependent Gaussian basis functions $\bm{\phi}_t$ for each DoF, and $\bm{\epsilon}_y \sim \mathcal{N}(\bm{0},\,\bm{\Sigma}_{\bm{y}})$ is the zero mean i.i.d. Gaussian noise with uncertainty $\bm{\Sigma}_{\bm{y}}$ (see~\cite{paraschos2018probabilistic}). 
	ProMPs employ a phase variable $z \in [0, 1]$ that decouples the demonstrations $\bm{\tau} = \{\bm{y}_t\}_{t = z_0}^{z_T}$ from the time instances, which in turn allows for temporal modulation.  
	Table~\ref{tab:notation} provides relevant notation used in this paper.
	
	ProMPs assume that each demonstration is characterized by a different weight vector $\bm{w}$, leading to a distribution $\mathcal{P}(\bm{w} ; \bm{\theta}) = \mathcal{N}(\bm{w}|\bm{\mu}_{\bm{w}}, \bm{\Sigma}_{\bm{w}})$.
	Consequently, the distribution of $\bm{y}_t$, $\mathcal{P}(\bm{y}_t;\bm{\theta})$ is
	\begin{equation} \label{eq3}
		\mathcal{P}(\bm{y}_t;\bm{\theta}) = \int \mathcal{N}(\bm{y}_t|\bm{\Psi}_t \bm{w}, \bm{\Sigma}_{\bm{y}}) \mathcal{N}(\bm{w}|\bm{\mu}_{\bm{w}}, \bm{\Sigma}_{\bm{w}}) d\bm{w}
		= \mathcal{N}(\bm{y}_t|\bm{\Psi}_t \bm{\mu}_{\bm{w}},\bm{\Psi}_t \bm{\Sigma}_{\bm{w}} \bm{\Psi}_t^\trsp+\bm{\Sigma}_{\bm{y}}) .
	\end{equation}

	\paragraph{Learning from demonstrations:}
	The learning process of ProMPs mainly consists of estimating the weight distribution $\mathcal{P}(\bm{w} ; \bm{\theta})$. To do so, a weight vector $\bm{w}_n$, representing the $n$-th demonstration as in~\eqref{eq:ClassicProMP}, is estimated by maximum likelihood, leading to the solution of the form 
	\begin{equation}\label{eq:weightProMP}
		\bm{w}_n = (\bm{\Psi}^{\trsp} \bm{\Psi} +\lambda \mathbf{I})^{-1}\bm{\Psi}^{\trsp} \bm{Y}_n ,
	\end{equation}
	where $\bm{Y}_n = \left[\bm{y}_{n,1}^{\trsp} \ldots \bm{y}_{n,T}^{\trsp}\right]^{\trsp}$ concatenates all the trajectory points, and $\mathbf{\Psi}$ consists of all the time instances for the basis-functions matrix $\mathbf{\Psi}_t$. Given a set of $N$ demonstrations, the weight distribution parameters $\bm{\theta} = \{\bm{\mu_w}, \bm{\Sigma_w} \}$ are estimated by maximum likelihood (see Algorithm~\ref{alg:alg1} in Appendix $1$). 
	
	\paragraph{Trajectory modulation:} 
	To adapt to new situations, ProMPs allow for trajectory modulation to via-points or target positions by conditioning the motion to reach a desired point $\bm{y}_t^*$ with associated covariance $\bm{\Sigma_y}^*$. This results into the conditional probability $\mathcal{P}(\bm{w}|\bm{y}_t^*)~\propto~\mathcal{N}(\bm{y}_t^*| \bm{\Psi}_t\bm{w},\bm{\Sigma_y}^*)\mathcal{N}(\bm{w}|\bm{\mu_w}, \bm{\Sigma_w})$, whose parameters can be computed as follows 
	\begin{equation}\label{eq:condProMP}
		\bm{\mu_w}^* = \bm{\Sigma_w}^* \left( \bm{\Psi}_t^\trsp \bm{\Sigma}_{\bm{y}}^{*^{-1}} \bm{y}_t^* + \bm{\Sigma_w}^{-1} \bm{\mu_w} \right) , \quad
		\bm{\Sigma_w}^* = \left( \bm{\Sigma_w}^{-1} + \bm{\Psi}_t^\trsp \bm{\Sigma}_{\bm{y}}^{*^{-1}} \bm{\Psi}_t \right)^{-1} .
	\end{equation}
	
	\paragraph{Blending:}
	\label{subsub:BlendingProMP}
	By computing a product of trajectory distributions, ProMPs blend different movement primitives into a single motion. The blended trajectory follows a distribution $\mathcal{P}(\bm{y}_t^{\tiny{+}})~=~\prod_{s=1}^{S} \mathcal{P}_s(\bm{y}_t)^{\alpha_{t,s}}$, for a set of $S$ ProMPs, where $\mathcal{P}_s(\bm{y}_t) = \mathcal{N}(\bm{y}_t|\bm{\mu}_{t,s}, \bm{\Sigma}_{t,s})$, with associated blending weight $\alpha_{t,s} \in \left[0,1\right]$. The parameters of $\mathcal{P}(\bm{y}_t^+) = \mathcal{N}(\bm{y}_t^+ | \bm{\mu}_t^{\tiny{+}}, \bm{\Sigma}_t^{\tiny{+}})$ are estimated as follows
	\begin{equation} \label{eq:PoGProMP}
		\bm{\Sigma}_t^{\tiny{+}} = \left( \sum_{s=1}^{S} \alpha_{t,s}\bm{\Sigma}_{t,s}^{-1} \right)^{-1}, \quad \mathrm{and} \quad 
		\bm{\mu}_t^{\tiny{+}} = \bm{\Sigma}_t^{\tiny{+}} \left(\sum_{s=1}^{S} \alpha_{t,s}\bm{\Sigma}_{t,s}^{-1} \bm{\mu}_{t,s}\right).				
	\end{equation}
	
	\paragraph{Task parametrization:}
	\label{subsub:TaskParamProMP}
	ProMPs also exploit task parameters to adapt the robot motion to, for example, target objects for reaching tasks. Formally, ProMPs consider an external state $\hat{\bm{s}}$ and learn an affine mapping from $\hat{\bm{s}}$ to the mean weight vector $\bm{\mu_w}$, leading to the joint distribution $\mathcal{P}(\bm{w},\hat{\bm{s}})~=~\mathcal{N}(\bm{w}|\bm{O} \hat{\bm{s}}+\bm{o},\bm{\Sigma_w})\mathcal{N}(\hat{\bm{s}}|\bm{\mu_{\hat{s}}},\bm{\Sigma_{\hat{s}}})$,
	where $\{\bm{O},\bm{o}\}$ are learned using linear regression.
	
	\subsection{Riemannian manifolds}
	\label{subsec:riemannian}
\begin{SCfigure}[50][tbp]
		\centering
		\includegraphics[width=.24\textwidth]{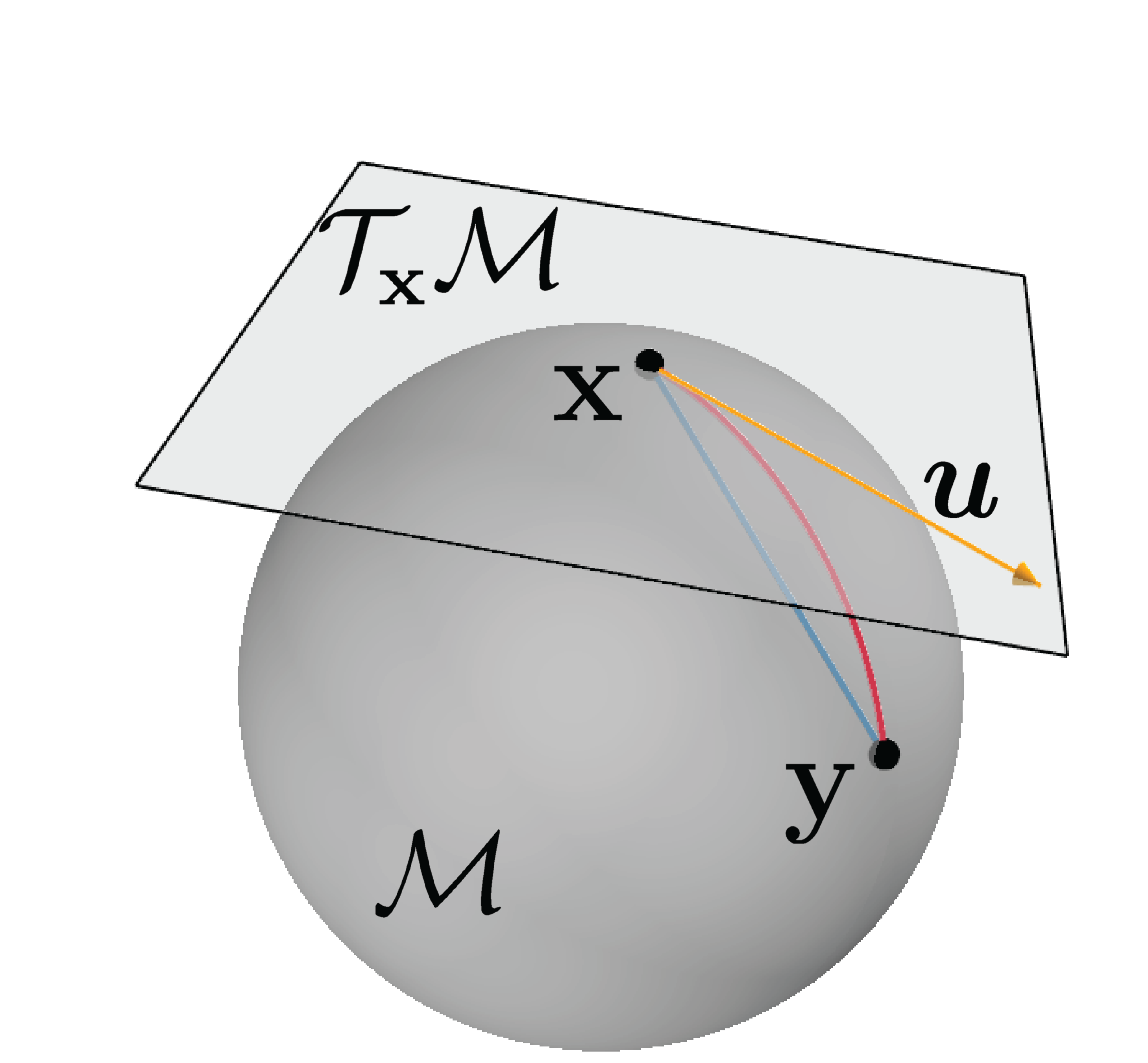}
		\includegraphics[trim={0cm 1.5cm 0cm 0cm},clip,width=.22\textwidth]{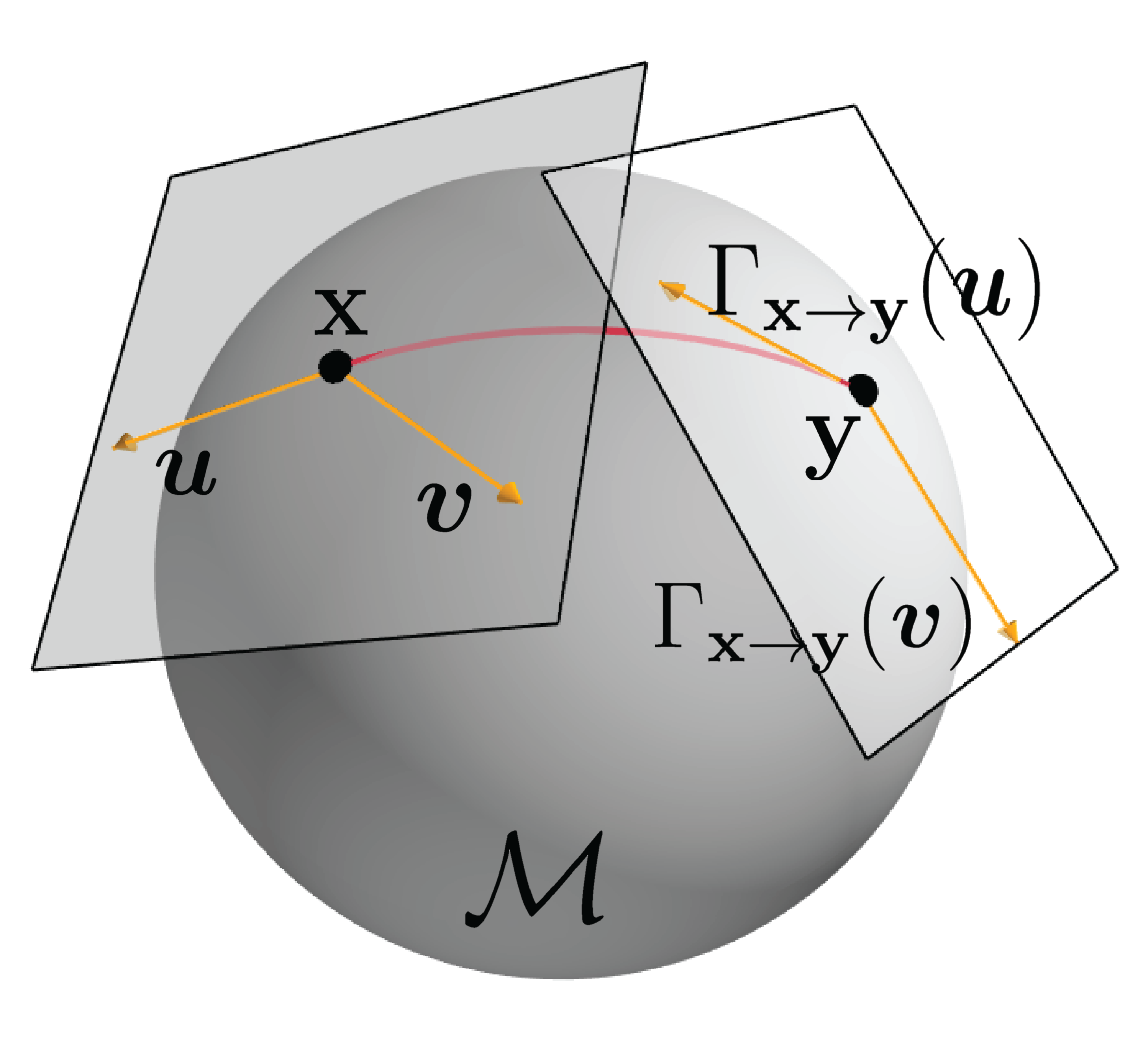}
		\caption{\emph{Left}: Points on the surface of the sphere $\mathcal{S}^2$, such as $\bm{\mathrm{x}}$ and $\bm{\mathrm{y}}$ belong to the manifold. The shortest path between $\bm{\mathrm{x}}$ and $\bm{\mathrm{y}}$ is the geodesic (\redGeod), which differs from the Euclidean path (\blueEucl). The vector $\bm{u}$ lies on the tangent space of $\bm{\mathrm{x}}$ such that $\bm{u} = \text{Log}_{\bm{\mathrm{x}}}(\bm{\mathrm{y}})$. \emph{Right}: $\Gamma_{\bm{\mathrm{x}}\rightarrow\bm{\mathrm{y}}}(\bm{u})$ and $\Gamma_{\bm{\mathrm{x}}\rightarrow\bm{\mathrm{y}}}(\bm{v})$ are the parallel transported vectors $\bm{u}$ and $\bm{v}$ from $\mathcal{T}_{\bm{\mathrm{x}}}\mathcal{M}$ to $\mathcal{T}_{\bm{\mathrm{y}}}\mathcal{M}$. The inner product between vectors is conserved by this operation.}
		\label{Fig:SphereManifold}
		\vspace{-0.1cm}
	\end{SCfigure}

	Since unit quaternions must satisfy a unit-norm constraint, they do not lie on a vector space, thus the use of traditional Euclidean methods for operating these variables is inadequate. We exploit Riemannian geometry to formulate ProMPs on quaternion space as presented in~\S~\ref{sec:quaternion_promps}. 
	Formally, a Riemannian manifold $\mathcal{M}$ is a $m$-dimensional topological space with a globally defined differential structure, where each point locally resembles an Euclidean space $\mathbb{R}^m$. 
	For each point $\bm{\mathrm{x}}\!\in\!\mathcal{M}$, there exists a tangent space $\mathcal{T}_{\bm{\mathrm{x}}} \mathcal{M}$ that is a vector space consisting of the tangent vectors of all the possible smooth curves passing through $\bm{\mathrm{x}}$. 
	A Riemannian manifold is equipped with a smoothly-varying positive definite inner product called a Riemannian metric, which permits to define curve lengths in $\mathcal{M}$. 
	These curves, called geodesics, are the generalization of straight lines on the Euclidean space to Riemannian manifolds, as they are minimum-length curves between two points in $\mathcal{M}$ (see Fig.~\ref{Fig:SphereManifold}). 
	
	We exploit the Euclidean tangent spaces to operate with Riemannian data. To do so, we need mappings back and forth between $\mathcal{T}_{\bm{\mathrm{x}}} \mathcal{M}$ and $\mathcal{M}$, which are the exponential and logarithmic maps.
	The exponential map $\text{Exp}_{\bm{\mathrm{x}}}: \mathcal{T}_{\bm{\mathrm{x}}} \mathcal{M} \to \mathcal{M}$ maps a point $\bm{u}$ in the tangent space of $\bm{\mathrm{x}}$ to a point $\bm{\mathrm{y}}$ on the manifold, so that it lies on the geodesic starting at $\bm{\mathrm{x}}$ in the direction $\bm{u}$, and such that the geodesic distance $d_{\mathcal{M}}$ between $\bm{\mathrm{x}}$ and $\bm{\mathrm{y}}$ equals the distance between $\bm{\mathrm{x}}$ and $\bm{u}$. The inverse operation is the logarithmic map $\text{Log}_{\bm{\mathrm{x}}}:  \mathcal{M}\to \mathcal{T}_{\bm{\mathrm{x}}}\mathcal{M}$.
	Another useful operation is the parallel transport $\Gamma_{\bm{\mathrm{x}}\rightarrow\bm{\mathrm{y}}}: \mathcal{T}_{\bm{\mathrm{x}}}\mathcal{M}\to\mathcal{T}_{\bm{\mathrm{y}}}\mathcal{M}$, which allows us to operate with manifold elements lying on different tangent spaces. 
	The parallel transport moves elements between tangent spaces such that the inner product between two elements in the tangent space remains constant (see~\cite{Lee18Riemann,boumal2020intromanifolds} for further details). 
	Finally, let us introduce a Riemannian Gaussian distribution of a random variable $\bm{\mathrm{x}} \in \mathcal{M}$ as
	\begin{equation}
		\mathcal{N}_{\mathcal{M}}(\bm{\mathrm{x}} | \bm{\mu}, \bm{\Sigma}) = \frac{1}{\sqrt{(2\pi)^d |\bm{\Sigma}|}} e^{-\frac{1}{2} \text{Log}_{\bm{\mu}}(\bm{\mathrm{x}})^\trsp \bm{\Sigma}^{-1} \text{Log}_{\bm{\mu}}(\bm{\mathrm{x}})} ,
	\end{equation}
	with mean $\bm{\mu} \in \mathcal{M}$, and covariance $\bm{\Sigma} \in \mathcal{T}_{\bm{\mu}}\mathcal{M}$. This Riemannian Gaussian corresponds to an approximated maximum-entropy distribution for Riemannian manifolds, as introduced by~\citet{Pennec06RiemannStats}. 
	Table~\ref{tab:SphereOperations} in Appendix~\ref{app:background} provides the different expressions for the Riemannian distance, exponential and logarithmic maps, and parallel transport operation for the sphere manifold $\mathcal{S}^m$.

	\subsection{Geodesic regression}
	\label{subsec:GeodesicReg}
	A geodesic regression model generalizes linear regression to Riemannian manifolds. The regression is defined as $\bm{\mathrm{y}} = \text{Exp}_{\tilde{\bm{\mathrm{y}}}}(\bm{\epsilon})$, with  $\tilde{\bm{\mathrm{y}}} = \text{Exp}_{\bm{\mathrm{p}}}(x\bm{u})$, where $\bm{\mathrm{y}} \in \mathcal{M}$ and $x \in \mathbb{R}$ are respectively the output and input variables, $\bm{\mathrm{p}} \in \mathcal{M}$ is a base point on the manifold, $\bm{u} \in \mathcal{T}_{\bm{\mathrm{p}}}\mathcal{M}$ is a tangent vector at $\bm{\mathrm{p}}$, and the error $\bm{\epsilon}$ is a random variable taking values in the tangent space at $\tilde{\bm{\mathrm{y}}}$. As an analogy to linear regression, one can interpret $(\bm{\mathrm{p}},\bm{u})$ as an intercept $\bm{\mathrm{p}}$ and a slope $\bm{u}$ (see~\cite{Fletcher2013GeodesicRA} for details). 
	Given a set of points $\{\bm{\mathrm{y}}_1,\ldots,\bm{\mathrm{y}}_T\} \in \mathcal{M}$ and $\{x_1,\ldots,x_T\} \in \mathbb{R}$, geodesic regression finds a geodesic curve $\gamma \in \mathcal{M}$ that best models the relationship between all the $T$ pairs $(x_i,\bm{\mathrm{y}}_i)$. To do so, we minimize the sum-of-squared Riemannian distances (i.e., errors) between the model estimates and the observations, that is, $E(\bm{\mathrm{p}},\bm{u}) = \frac{1}{2} \sum_{i=1}^{T}{d_{\mathcal{M}}(\hat{\bm{\mathrm{y}}}_i,\bm{\mathrm{y}}_i)^2}$, where $\hat{\bm{\mathrm{y}}}_i = \text{Exp}_{\bm{\mathrm{p}}}(x_i\bm{u})$ is the model estimate on $\mathcal{M}$, $d_{\mathcal{M}}(\hat{\bm{\mathrm{y}}}_i,\bm{\mathrm{y}}_i) = \|\text{Log}_{\hat{\bm{\mathrm{y}}}_i}(\bm{\mathrm{y}}_i) \|$ is the Riemannian error, and the pair $(\bm{\mathrm{p}},\bm{u}) \in \mathcal{TM}$ is an element of the tangent bundle $\mathcal{TM}$. We can formulate a least-squares estimator of the geodesic model as a minimizer of such sum-of-squared Riemannian distances, i.e.,
	\begin{equation}\label{eq:GeodesicRegModel}
		(\hat{\bm{\mathrm{p}}},\hat{\bm{u}}) = \argmin_{(\bm{\mathrm{p}},\bm{u}) \in \mathcal{TM}} \frac{1}{2} \sum_{i=1}^{T}{d_\mathcal{M}(\hat{\bm{\mathrm{y}}}_i,\bm{\mathrm{y}}_i)^2} .
	\end{equation}
	
	The problem in \eqref{eq:GeodesicRegModel} does not yield an analytical solution like \eqref{eq:weightProMP}. As explained by~\citet{Fletcher2013GeodesicRA}, a solution can be obtained via gradient descent on Riemannian manifolds. 
	Note that this geodesic model considers only a scalar independent variable $x \in \mathbb{R}$, meaning that the derivatives are obtained along a \emph{single} geodesic curve parametrized by a \emph{single} tangent vector $\bm{u}$. 
	The extension to multivariate cases proposed by~\citet{MGML2014}, where $\bm{x} \in \mathbb{R}^M$, requires a slightly different approach to identify multiple geodesic curves (viewed as ``basis'' vectors in Euclidean space). 
	Multivariate general linear models on Riemannian manifolds (MGLM)~\cite{MGML2014} provides a solution to this problem.
	MLGM uses a geodesic basis $\bm{U} = \left[\bm{u}_1 \ldots \bm{u}_M\right]$ formed by multiple tangent vectors $\bm{u}_m \in \mathcal{T}_{\bm{\mathrm{p}}}\mathcal{M}$ of dimensionality $d = \operatorname{dim}(\mathcal{T}_{\bm{\mathrm{p}}}\mathcal{M})$, one for each dimension of $\bm{x}$. Then, the problem~\eqref{eq:GeodesicRegModel} can be reformulated as 
	\begin{equation} \label{eq:MultiGeodesicReg}
		(\hat{\bm{\mathrm{p}}},\hat{\bm{u}}_m) = \argmin_{(\bm{\mathrm{p}},\bm{u}_m) \in \mathcal{TM} \, \forall m} \; \frac{1}{2} \sum_{i=1}^{T}{d_\mathcal{M}(\hat{\bm{\mathrm{y}}}_i,\bm{\mathrm{y}}_i)^2} , \quad \text{with} \;  \hat{\bm{\mathrm{y}}}_i = \text{Exp}_{\bm{\mathrm{p}}}(\bm{U}\bm{x}_i) .
	\end{equation}
	This multivariate framework allows us to compute the weight vector, analogous to~\eqref{eq:weightProMP}, for a demonstration lying on a Riemannian manifold $\mathcal{M}$, for example $\mathcal{M} \equiv \mathcal{S}^3$. 
	

	\section{Orientation ProMPs}
	\label{sec:quaternion_promps}
	When human demonstrations involve Cartesian motion patterns (via kinesthetic teaching or teleoperation), it is necessary to have a learning model that encapsulates both translation and rotation movements of the robot end-effector. 
	This means that a demonstration trajectory $\bm{\tau} = \{\bm{y}_t\}_{t=1}^T$ is now composed of datapoints $\bm{y}_t \in \mathbb{R}^3 \times \mathcal{S}^3$, representing the full Cartesian pose of the end-effector at time step $t$. 
	In this case, the challenge is learning a ProMP in the orientation space, as the Euclidean case in $\mathbb{R}^3$ follows the classic ProMP introduced in~\S~\ref{subsec:promps}. 
	Therefore, we focus on how to extend ProMP to learn trajectories on $\mathcal{M} = \mathcal{S}^3$. 
	First of all, let us introduce an equivalent expression for $\hat{\bm{\mathrm{y}}}_i$, in the MGLM framework, such that it resembles the linear basis-function model in~\eqref{eq:ClassicProMP}. Specifically, the estimate $\hat{\bm{\mathrm{y}}}_i = \text{Exp}_{\bm{\mathrm{p}}}(\bm{U}\bm{x}_i) \equiv \text{Exp}_{\bm{\mathrm{p}}}(\bm{X}_i\bm{u})$ with $\bm{U} \in \mathbb{R}^{d \times M}, \bm{x}_i \in \mathbb{R}^M$, $\bm{X}_i = \text{blockdiag} (\bm{x}_1^\trsp, \ldots, \bm{x}_M^\trsp) \in \mathbb{R}^{d\times Md}$ and $\bm{u} = [\bm{u}_1^\trsp \ldots \bm{u}_M^\trsp]^\trsp \in \mathbb{R}^{Md}$. 

	This equivalence proves useful when establishing analogies between the classic ProMPs formulation and our approach for orientation trajectories.   
	
	Similarly to \eqref{eq:ClassicProMP}, a point $\bm{\mathrm{y}}_t \in \mathcal{M}$ of $\bm{\tau}$ can be represented as a \emph{geodesic} basis-function model 
	\begin{equation} \label{eq:OrientationProMP}
		\mathcal{P}(\bm{\mathrm{y}}_t | \bm{w}) = \mathcal{N}_{\tiny{\mathcal{M}}}(\bm{\mathrm{y}}_t | \text{Exp}_{\bm{\mathrm{p}}}(\bm{\Psi}_t\bm{w}), \bm{\Sigma}_{\bm{\mathrm{y}}}) , \quad \text{with} \quad \bm{\mu_{\mathrm{y}}} 
		= \text{Exp}_{\bm{\mathrm{p}}}(\bm{\Psi}_t\bm{w}) \in \mathcal{M},
	\end{equation} 
	where $\bm{\mathrm{p}}$ is a fixed base point on $\mathcal{M}$, $\bm{w} = \left[\bm{w}_1^\trsp \ldots \bm{w}_{N_\phi}^\trsp\right]^\trsp$ is a large weight vector concatenating $N_\phi$ weight vectors $\bm{w}_n \in \mathcal{T}_{\bm{\mathrm{p}}}\mathcal{M}$, $\bm{\Psi}_t$ is the same matrix of time-dependent basis functions as in~\eqref{eq:ClassicProMP}, and $\bm{\Sigma}_{\bm{\mathrm{y}}}$ is a covariance matrix encoding the uncertainty on $ \mathcal{T}_{\bm{\mu}_{\bm{\mathrm{y}}}}\mathcal{M}$. Two specific aspects about this formulation deserve special attention:  (\emph{i})~the mean $\bm{\mu_{\mathrm{y}}}$ of the Riemannian Gaussian distribution in~\eqref{eq:OrientationProMP}, exploits the aforementioned equivalent formulation of MGLM; and (\emph{ii})~the weight vectors forming $\bm{w}$ in~\eqref{eq:OrientationProMP} correspond to the vector composing the geodesic basis of MGLM. 
	
	As a demonstration is characterized by a different weight vector $\bm{w}$, again we can compute a distribution $\mathcal{P}(\bm{w} ; \bm{\theta}) = \mathcal{N}(\bm{w}|\bm{\mu}_{\bm{w}}, \bm{\Sigma}_{\bm{w}})$. Therefore, the marginal distribution of $\bm{\mathrm{y}}_t$ can be defined as
	\begin{equation} \label{eq:MarginalDistributionS3}
		\mathcal{P}(\bm{\mathrm{y}};\bm{\theta}) = \int \mathcal{N}_{\tiny{\mathcal{M}}}(\bm{\mathrm{y}} | \text{Exp}_{\bm{\mathrm{p}}}(\bm{\Psi}\bm{w}), \bm{\Sigma}_{\bm{\mathrm{y}}}) \mathcal{N}(\bm{w}|\bm{\mu}_{\bm{w}}, \bm{\Sigma}_{\bm{w}}) d\bm{w} , \;\; \text{with} \;\; \bm{\mu_{\mathrm{y}}} = \text{Exp}_{\bm{\mathrm{p}}}(\bm{\Psi}\bm{w}) \in \mathcal{M},
	\end{equation}
	which depends on two probability distributions that lie on different manifolds. Using the Riemannian operations described in~\S~\ref{sec:background}, we obtain the final marginal (see Appendix~\ref{app:quaternion_promps_marg})
	\begin{equation}\label{eq:FinalMarginalOriProMP}
		\mathcal{P}(\bm{\mathrm{y}};\bm{\theta}) = \mathcal{N}_{\tiny{\mathcal{M}}}(\bm{\mathrm{y}} | \hat{\bm{\mu}}_{\bm{\mathrm{y}}}, \hat{\bm{\Sigma}}_{\bm{\mathrm{y}}}) , \; \text{with} \; \; \hat{\bm{\mu}}_{\bm{\mathrm{y}}}=\text{Exp}_{\bm{\mathrm{p}}}(\bm{\Psi}\bm{\mu}_{\bm{w}}),  \hat{\bm{\Sigma}}_{\bm{\mathrm{y}}} = \Gamma_{\bm{\mathrm{p}} \rightarrow \hat{\bm{\mu}}_{\bm{\mathrm{y}}}}(\bm{\Psi} \bm{\Sigma}_{\bm{w}}\bm{\Psi}^\trsp + \tilde{\bm{\Sigma}}_{\bm{\mathrm{y}}}) ,
	\end{equation}
	where $\tilde{\bm{\Sigma}}_{\bm{\mathrm{y}}}$ and $\hat{\bm{\Sigma}}_{\bm{\mathrm{y}}}$ are parallel-transported covariances of the geodesic model~\eqref{eq:OrientationProMP} and the final marginal, respectively.

\begin{SCfigure}[50][tbp]
		\centering
		\includegraphics[trim={8cm 3cm 1cm 7cm},clip,width=.3\textwidth]{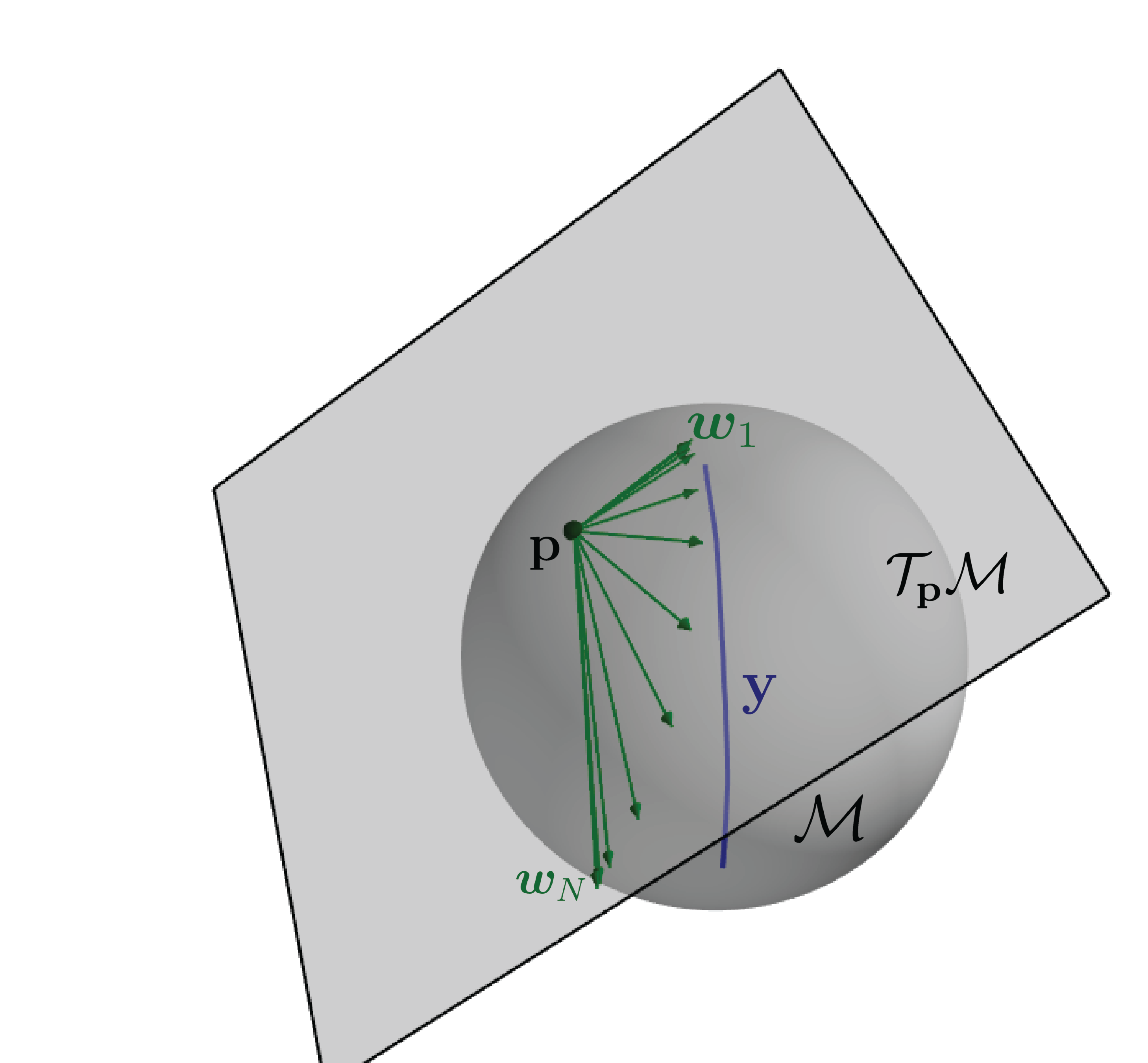}
		\caption{Illustration of multivariate general linear regression on the sphere manifold $\mathcal{S}^2$ used to learn the weights of orientation ProMPs. Given the trajectory $\bm{\mathrm{y}}$, the origin $\bm{\mathrm{p}}$ of the tangent space $\mathcal{T}_{\bm{\mathrm{p}}}\mathcal{M}$, and the tangent weight vectors $\bm{w}_m$ are estimated via~\eqref{eq:weightOriProMP}.}
		\label{Fig:GeodesicReg}
		\vspace{-.3cm}
	\end{SCfigure}

	\paragraph{Learning from demonstrations via MGLM:}
	For each demonstration $n$, we estimate a weight vector $\hat{\bm{w}}_n = \left[\hat{\bm{w}}_1^\trsp \ldots \hat{\bm{w}}_{N_{\phi}}^\trsp\right]^\trsp$ using MGLM (illustrated in Fig.~\ref{Fig:GeodesicReg}). Firstly, we resort to the equivalent expression for $\bm{\mathrm{y}}_t$ introduced previously, where $\text{Exp}_{\bm{\mathrm{p}}}(\bm{W}\bm{\phi}_t) \equiv \text{Exp}_{\bm{\mathrm{p}}}(\bm{\Psi}_t\bm{w})$, with $\bm{W} = \left[\bm{w}_1 \ldots \bm{w}_{N_{\phi}}\right]$ and $N_{\phi}$ being the number of basis functions. Secondly, we consider a demonstrated quaternion trajectory $\bm{\tau}_n = \{\bm{\mathrm{y}}_t\}_{t=1}^T$ with $\bm{\mathrm{y}}_t \in \mathcal{S}^3$. Then, analogous to~\eqref{eq:weightProMP} in Euclidean space, $\hat{\bm{w}}_n$ is estimated by leveraging~\eqref{eq:MultiGeodesicReg}, leading to
	\begin{equation}\label{eq:weightOriProMP}
		(\hat{\bm{\mathrm{p}}},\hat{\bm{w}}_m) = \argmin_{(\bm{\mathrm{p}},\bm{w}_m) \in \mathcal{TM} \, \forall m} \; E(\bm{\mathrm{p}}, \bm{w}_m) , \quad \text{with} \quad  E(\bm{\mathrm{p}}, \bm{w}_m) = \frac{1}{2} \sum_{t=1}^{T}{d_\mathcal{M}(\text{Exp}_{\bm{\mathrm{p}}}(\bm{W}\bm{\phi}_t), \bm{\mathrm{y}}_t)^2} ,
	\end{equation}
	where $\bm{\phi}_t$ is the vector of Gaussian basis functions at time $t$, and $\bm{W}$ contains the set of estimated tangent weight vectors $\hat{\bm{w}}_m \in \mathcal{T}_{\hat{\bm{\mathrm{p}}}}\mathcal{M}$ (i.e., $N_\phi$ tangent vectors emerging out from the point $\bm{\mathrm{p}} \in \mathcal{M}$). 
	To solve~\eqref{eq:weightOriProMP}, we need to compute the gradients of $E(\bm{\mathrm{p}}, \bm{w}_m)$ with respect to $\bm{\mathrm{p}}$ and each $\bm{w}_m$. As explained in Appendix~$2.2$, these gradients depend on the so-called adjoint operators, which broadly speaking, bring each error term $\text{Log}_{\hat{\bm{\mathrm{y}}}_t}(\bm{\mathrm{y}}_t)$ from $\mathcal{T}_{\hat{\bm{\mathrm{y}}}_t}\mathcal{M}$ to $\mathcal{T}_{\bm{\mathrm{p}}}\mathcal{M}$, with $\hat{\bm{\mathrm{y}}}_t = \text{Exp}_{\bm{\mathrm{p}}}(\bm{W}\bm{\phi}_t)$. Therefore, these adjoint operators can be approximated as parallel transport operations as proposed in~\cite{MGML2014}. This leads to the following reformulation of the error function of~\eqref{eq:weightOriProMP} 
	\begin{equation}\label{eq:ParallelTransportError}
		E(\bm{\mathrm{p}},\bm{w}_m) = \frac{1}{2} \sum_{t=1}^{T}\| \Gamma_{\hat{\bm{\mathrm{y}}}_t \rightarrow \bm{\mathrm{p}}}( \text{Log}_{\hat{\bm{\mathrm{y}}}_t}(\bm{\mathrm{y}}_t)) \|^2 .
	\end{equation}
	
	With the gradients of \eqref{eq:ParallelTransportError} (given in Appendix~\ref{subapp:mglm}), we can now estimate both the vector $\bm{\mathrm{p}}_n$ and the weight matrix $\bm{W}_n$, for each demonstration $n$. Note that each demonstration may lead to different estimates of $\bm{\mathrm{p}}$, which defines the \emph{base point} on $\mathcal{M}$ used to estimate $\bm{w}_m \in \mathcal{T}_{\bm{\mathrm{p}}}\mathcal{M}$. This may produce different tangent spaces across demonstrations, and therefore diverse tangent weight vectors. An effective way to overcome this is to assume that all demonstrations share the same tangent space base $\bm{\mathrm{p}}$, which is the same assumption made when defining the geodesic basis-function model~\eqref{eq:OrientationProMP}. So, we only need to estimate $\bm{\mathrm{p}}$ for a single demonstration, and use it to estimate all tangent weight vectors for the whole set of demonstrations. Then, given a set of $N$ demonstrations, the weight distribution parameters $\bm{\theta} = \{\bm{\mu}_{\bm{w}}, \bm{\Sigma}_{\bm{w}} \}$ are estimated by standard maximum likelihood as $\bm{w}_n~\in~\mathcal{T}_{\bm{\mathrm{p}}}\mathcal{M}~=~\mathbb{R}^3 \subset \mathbb{R}^4$. The complete learning process for orientation ProMPs is given in Algorithm~\ref{alg:alg2} of the appendix.

	\paragraph{Trajectory modulation:}
	\label{subsec:condOriProMP}
	We formulate a trajectory modulation technique (i.e., to adapt to new situations) by conditioning the motion to reach a desired trajectory point $\bm{\mathrm{y}}_t^* \in \mathcal{M}$ with associated covariance $\bm{\Sigma}_{\bm{\mathrm{y}}}^* \in \mathcal{T}_{\bm{\mathrm{y}}_t^*}\mathcal{M}$. This results into the conditional probability $\mathcal{P}(\bm{w}|\bm{\mathrm{y}}_t^*)~\propto~\mathcal{N}_{\tiny{\mathcal{M}}}(\bm{\mathrm{y}}_t^*| \text{Exp}_{\bm{\mathrm{p}}}(\bm{\Psi}_t\bm{w}),\bm{\Sigma}_{\bm{\mathrm{y}}}^*)\mathcal{N}(\bm{w}|\bm{\mu}_{\bm{w}}, \bm{\Sigma}_{\bm{w}})$, which depends on two probability distributions that lie on different manifolds, similarly to~\eqref{eq:MarginalDistributionS3}. We leverage the fact that the mean 
	$\bm{\mu_{\mathrm{y}}}$ depends on $\bm{\mathrm{p}} \in \mathcal{M}$, which is the basis of $\mathcal{T}_{\bm{\mathrm{p}}}\mathcal{M}$ where the weight distribution lies on. Thus, we rewrite the conditional distribution as follows
	\begin{equation}\label{eq:ConditionalOriProMP}
		\mathcal{P}(\bm{w}|\text{Log}_{\bm{\mathrm{p}}}(\bm{\mathrm{y}}_t^*))~\propto~\mathcal{N}(\text{Log}_{\bm{\mathrm{p}}}(\bm{\mathrm{y}}_t^*)| \bm{\Psi}_t\bm{w},\tilde{\bm{\Sigma}}_{\bm{\mathrm{y}}}^*)\mathcal{N}(\bm{w}|\bm{\mu}_{\bm{w}}, \bm{\Sigma}_{\bm{w}})
		= \mathcal{N}(\bm{w}|\bm{\mu_w}^*, \bm{\Sigma_w}^*) ,
	\end{equation}
	where $\tilde{\bm{\Sigma}}_{\bm{\mathrm{y}}}^* = \Gamma_{\bm{\mathrm{y}}_t^* \rightarrow \bm{\mathrm{p}}}(\bm{\Sigma}_{\bm{\mathrm{y}}}^*)$, and $\{\bm{\mu}_{\bm{w}}^*, \bm{\Sigma}_{\bm{w}}^*\}$ are the parameters to estimate for the resulting conditional distribution. Since both distributions now lie on $\mathcal{T}_{\bm{\mathrm{p}}}\mathcal{M}$, which is embedded in the Euclidean space, we can estimate $\{\bm{\mu}_{\bm{w}}^*, \bm{\Sigma}_{\bm{w}}^*\}$ similarly to the classic ProMP conditioning procedure, with special care of parallel-transporting the covariance matrices. So, the updated parameters are
	\begin{equation}\label{eq:condOriProMP}
		\bm{\mu_w}^* = \bm{\Sigma_w}^* \left( \bm{\Psi}_t^\trsp \tilde{\bm{\Sigma}}_{\bm{\mathrm{y}}}^{*^{-1}} \text{Log}_{\bm{\mathrm{p}}} (\bm{\mathrm{y}}_t^*) + \bm{\Sigma_w}^{-1} \bm{\mu_w} \right) ,\quad \mathrm{and} \quad
		\bm{\Sigma_w}^* = \left( \bm{\Sigma_w}^{-1} + \bm{\Psi}_t^\trsp \tilde{\bm{\Sigma}}_{\bm{\mathrm{y}}}^{*^{-1}} \bm{\Psi}_t \right)^{-1} .
	\end{equation}
	From the new weight distribution, we can obtain a new marginal distribution $\mathcal{P}(\bm{\mathrm{y}};\bm{\theta}^*)$ via~\eqref{eq:FinalMarginalOriProMP}. 
	
	\paragraph{Blending:}
	\label{subsec:blendingOriProMP}
	When it comes to blend motion primitives in $\mathcal{M}$, one needs to consider that each of them is parametrized by a set of weight vectors lying on different tangent spaces $\mathcal{T}_{\bm{\mathrm{p}}}\mathcal{M}$. Therefore, the weighted product of Gaussian distributions needs to be reformulated. To do so, we resort to the Gaussian product formulation on Riemannian manifolds introduced by~\citet{Zeestraten17riemannian}, where the log-likelihood of the product is iteratively maximized using a gradient-based approach as proposed in~\cite{Dubbelman11}. We here provide the iterative updates of the blended distribution $\mathcal{P}(\bm{y}^+) = \mathcal{N}_{\mathcal{M}}(\bm{y}^+ | \bm{\mu}^{\tiny{+}}, \bm{\Sigma}^{\tiny{+}})$, while the full solution is given in Appendix~\ref{subapp:blendingOriProMP}. The iterative estimation of $\bm{\mu}^+$ is
	\begin{equation}
		\Delta_{\bm{\mu}^+_k} = \left( \sum_{s=1}^{S} \alpha_{s}\bm{\Lambda}_{\bm{\mathrm{y}},s} \right)^{-1} \left(\sum_{s=1}^{S} \alpha_{s}\bm{\Lambda}_{\bm{\mathrm{y}},s} \text{Log}_{\bm{\mu}^+_{k}}(\bm{\mu}_{\bm{\mathrm{y}},s})\right) , \quad \text{and} \quad				
		\bm{\mu}^+_{k+1}  \leftarrow \text{Exp}_{\bm{\mu}^+_{k}}(\Delta_{\bm{\mu}^+_k}) , 
	\end{equation}
	for a set of $S$ skills, where $\bm{\Lambda}_{\bm{\mathrm{y}}, s} = \Gamma_{\bm{\mu}_{\bm{\mathrm{y}}, s} \rightarrow \bm{\mu}^+_k}(\bm{\Sigma}_{\bm{\mathrm{y}}, s}^{-1})$, and $\alpha_{s}$ is the blending weight associated to the skill $s$. After convergence at iteration $K$, we obtain the final parameters as follows
	\begin{equation}
		\bm{\mu}^+ \leftarrow \bm{\mu}^+_K \quad \text{and} \quad \bm{\Sigma}^+ = \left(\sum_{s=1}^{S} \alpha_{s}\bm{\Lambda}_{\bm{\mathrm{y}},s} \right)^{-1} .
	\end{equation}
	
	\paragraph{Task parametrization:}
	Classic ProMP allows for adapting the weight distribution $\mathcal{P}(\bm{w} ; \bm{\theta}) = \mathcal{N}(\bm{w}|\bm{\mu}_{\bm{w}}, \bm{\Sigma}_{\bm{w}})$ as a function of an external task parameter $\hat{\bm{s}}$, as explained in Section~\ref{subsub:TaskParamProMP}. 
	This task parametrization straightforwardly applies to our method as the weight vectors $\bm{w}_n \in \mathcal{T}_{\bm{\mathrm{p}}}\mathcal{M} \subset \mathbb{R}^4$,  as long as the task parameter $\hat{\bm{s}}$ is Euclidean. 
	However, if $\hat{\bm{s}} \in \mathcal{M}$, we can learn a joint probability distribution $\mathcal{P}(\bm{w}, \hat{\bm{s}})$ using a Gaussian mixture model on Riemannian manifolds as proposed in~\cite{Zeestraten17riemannian,Jaquier21}. Subsequently, we can employ Gaussian mixture regression to compute $\mathcal{P}(\bm{w} | \hat{\bm{s}}^*)$ during reproduction when a new task parameter $\hat{\bm{s}}^*$ is provided. We refer the reader to the works of~\citet{Zeestraten17riemannian} and \citet{Jaquier21} for details on how to compute the distributions $\mathcal{P}(\bm{w}, \hat{\bm{s}})$ and $\mathcal{P}(\bm{w} | \hat{\bm{s}}^*)$. 
	

	\section{Experiments}
	\label{sec:experiments}
	
	\paragraph{Synthetic data on $\mathcal{S}^2$:}
	To illustrate how model learning, trajectory reproduction, via-point adaptation, and skills blending work in our approach, we used a dataset of hand-written letters. 
	The original trajectories were generated in $\mathbb{R}^2$ and later projected to $\mathcal{S}^2$ by a unit-norm mapping. 
	Each letter in the dataset was demonstrated $N=8$ times, and a simple smoothing filter was applied for visualization purposes. 
	We trained $4$ ProMPs, one for each letter in the dataset $\{\mathsf{G}, \mathsf{I}, \mathsf{J}, \mathsf{S}\}$. 
	All models were trained following Algorithm~\ref{alg:alg2} with hyperparameters given in Appendix~\ref{app:experiments}.   
	Figure~\ref{Fig:MarginalCondToy} shows the demonstration data, marginal distribution $\mathcal{P}(\bm{\mathrm{y}};\bm{\theta})$ computed via~\eqref{eq:MarginalDistributionS3}, and via-point adaptation obtained from~\eqref{eq:ConditionalOriProMP} and~\eqref{eq:condOriProMP}, corresponding to the $\mathsf{G}$ and $\mathsf{S}$ models. 
	The mean of $\mathcal{P}(\bm{\mathrm{y}};\bm{\theta})$ follows the demonstrations pattern, and the covariance profile captures the demonstrations variability in $\mathcal{S}^2$. 
	Note that the trajectories $\mathsf{G}$ and $\mathsf{S}$ display very elaborated ``motion" patterns that might be more complex than those observed in realistic robotic settings.
	Concerning the via-point adaptation, we chose a random point $\bm{\mathrm{y}}^* \in \mathcal{S}^2$ with associated $\bm{\Sigma}_{\bm{\mathrm{y}}}^* = \num{e-3}\bm{I}$ (i.e., high precision while passing through $\bm{\mathrm{y}}^*$). 
	As shown in Fig.~\ref{Fig:MarginalCondToy}, our approach smoothly adapts the trajectory and its covariance while accurately passing through $\bm{\mathrm{y}}^*$.        
	
	\begin{figure*}[t]
		\centering
		\includegraphics[width=.245\textwidth]{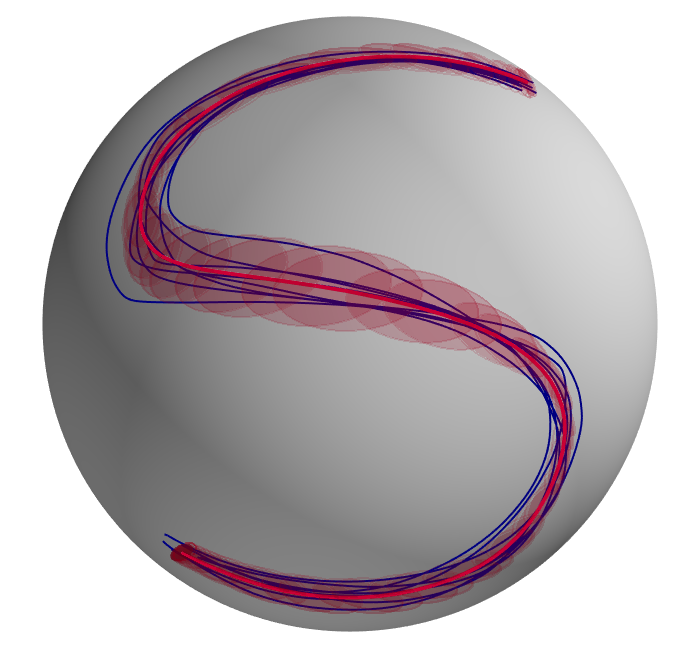}
		\includegraphics[width=.245\textwidth]{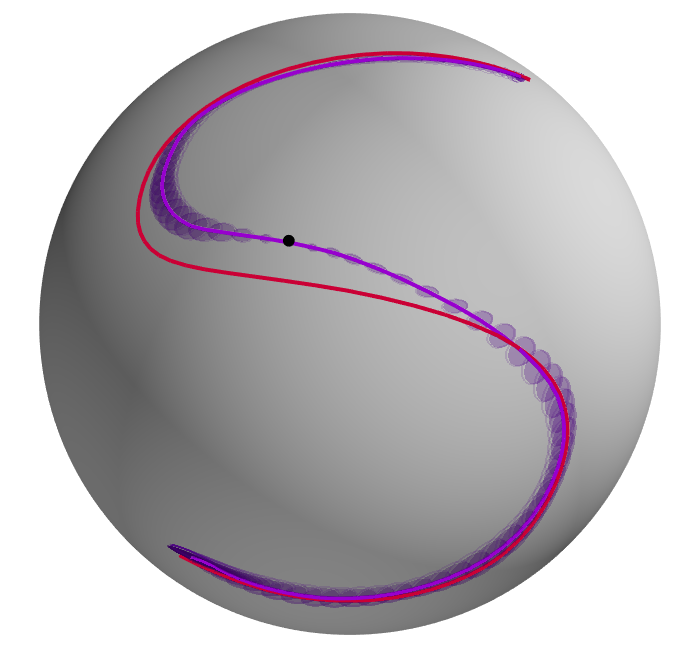}
		\includegraphics[width=.245\textwidth]{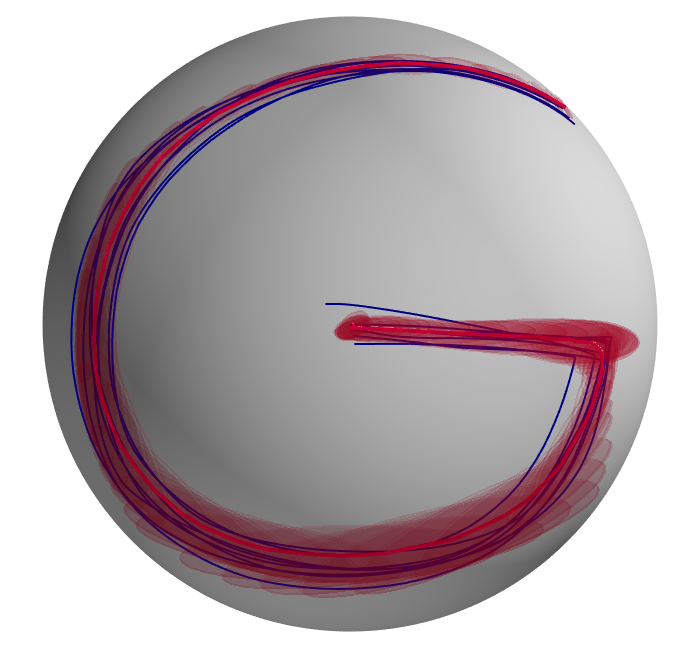}
		\includegraphics[width=.245\textwidth]{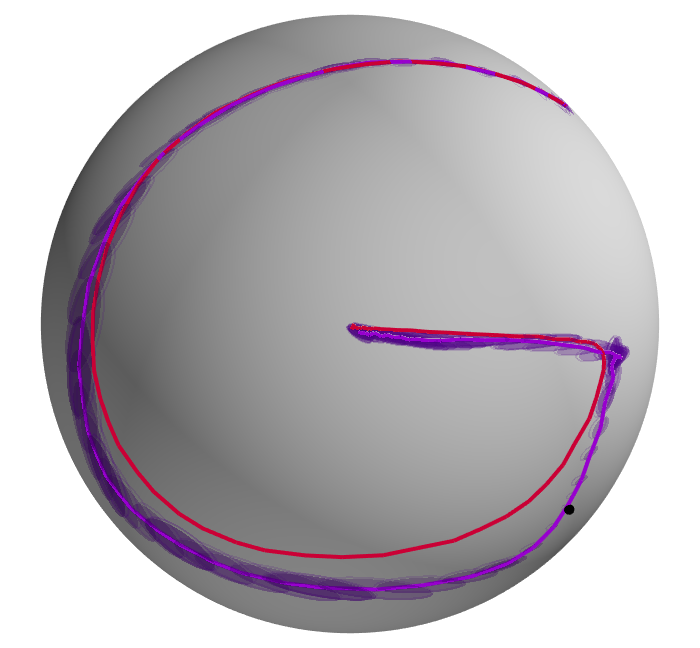}
		\caption{Demonstrated trajectories (\bluedemo) in $\mathcal{S}^2$, marginal distribution $\mathcal{P}(\bm{\mathrm{y}};\bm{\theta})$ (mean \redrepro, covariance \lightredellipse), and via-point adaptation (mean \purplerepro, covariance \purpleellipse, and via-point \blackcircle) for models trained over $\mathsf{S}$ and $\mathsf{G}$ datasets.}
		\label{Fig:MarginalCondToy}
		\vspace{-0.35cm}
	\end{figure*}
	
	\begin{SCfigure}[50][t]
		\centering
		\includegraphics[width=.24\textwidth]{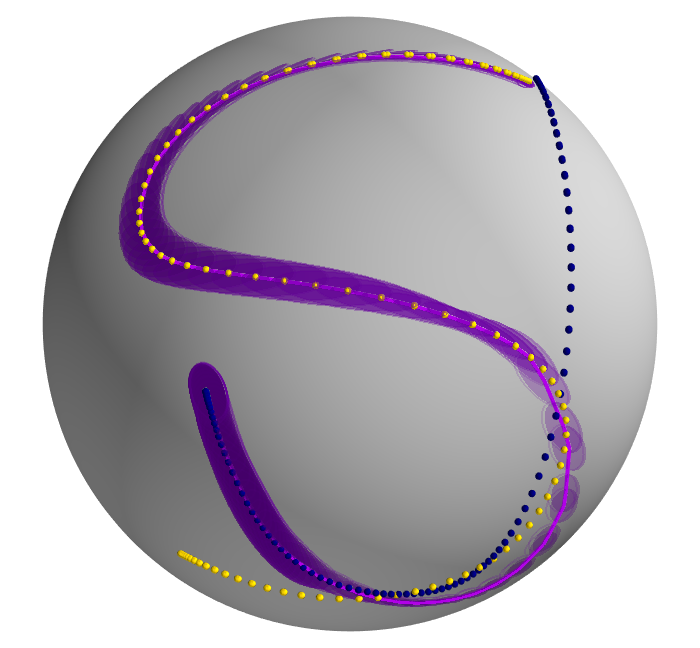}
		\includegraphics[width=.24\textwidth]{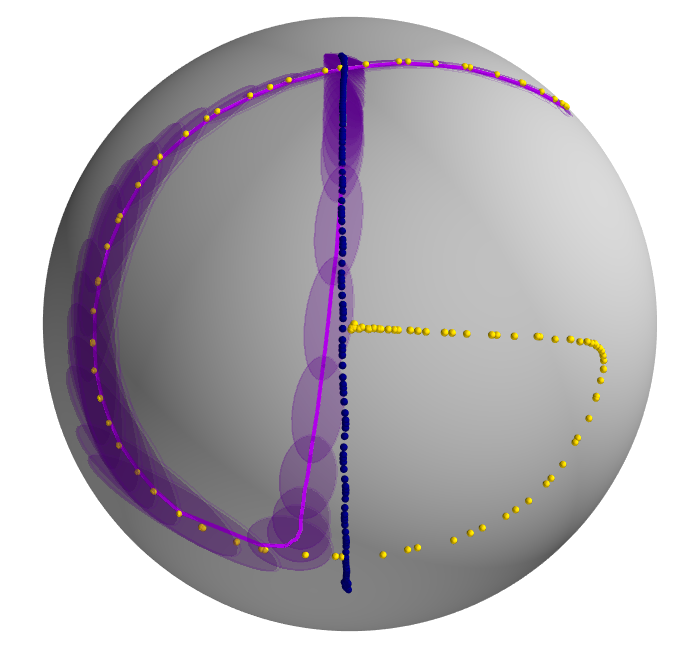}
		\caption{Trajectory distribution blending via orientation ProMP for datasets $\{\mathsf{S}, \mathsf{J}\}$ and $\{\mathsf{G}, \mathsf{I}\}$. The resulting (blended) distribution trajectory (mean \purplerepro, and covariance \purpleellipse) starts from the first letter in the dataset (\yellowcircle) and then smoothly joins the trajectory of the second letter (\bluecircle).}
		\label{Fig:BlendingToy}
		\vspace{-0.35cm}
	\end{SCfigure}
	
	To test the blending process, we used the following subsets of our original dataset: $\{\mathsf{G}, \mathsf{I}\}$ and $\{\mathsf{S}, \mathsf{J}\}$. 
	The goal was to generate a trajectory starting from the first letter in the set, and then smoothly switching midway to the second letter. 
	Figure~\ref{Fig:BlendingToy} shows the resulting blended trajectories for the two aforementioned cases, where our approach smoothly blends the two given trajectory distributions by following the procedure introduced in \S~\ref{subsec:blendingOriProMP}. 
	Note that the blending behavior strongly depends on the temporal evolution of the weights $\alpha_{s} \in [0, 1]$ associated to each skill $s$.
	We used a sigmoid-like function for the weights $\alpha_{s}^{(\mathsf{I})}$ and $\alpha_{s}^{(\mathsf{J})}$, while  $\alpha_{s}^{(\mathsf{G})} = 1 - \alpha_{s}^{(\mathsf{I})}$ and $\alpha_{s}^{(\mathsf{S})} = 1 - \alpha_{s}^{(\mathsf{J})}$. 
	The foregoing results show that our approach successfully learns and reproduces trajectory distributions on $\mathcal{S}^2$, and provides full via-point adaptation and blending capabilities. 
	We now turn our attention to robotic experiments where we learn and synthesize full-pose movement primitives. 
	
\begin{figure}[t]
		\centering
		\includegraphics[trim={4cm 5cm 32cm 2cm},clip,width=.155\textwidth]{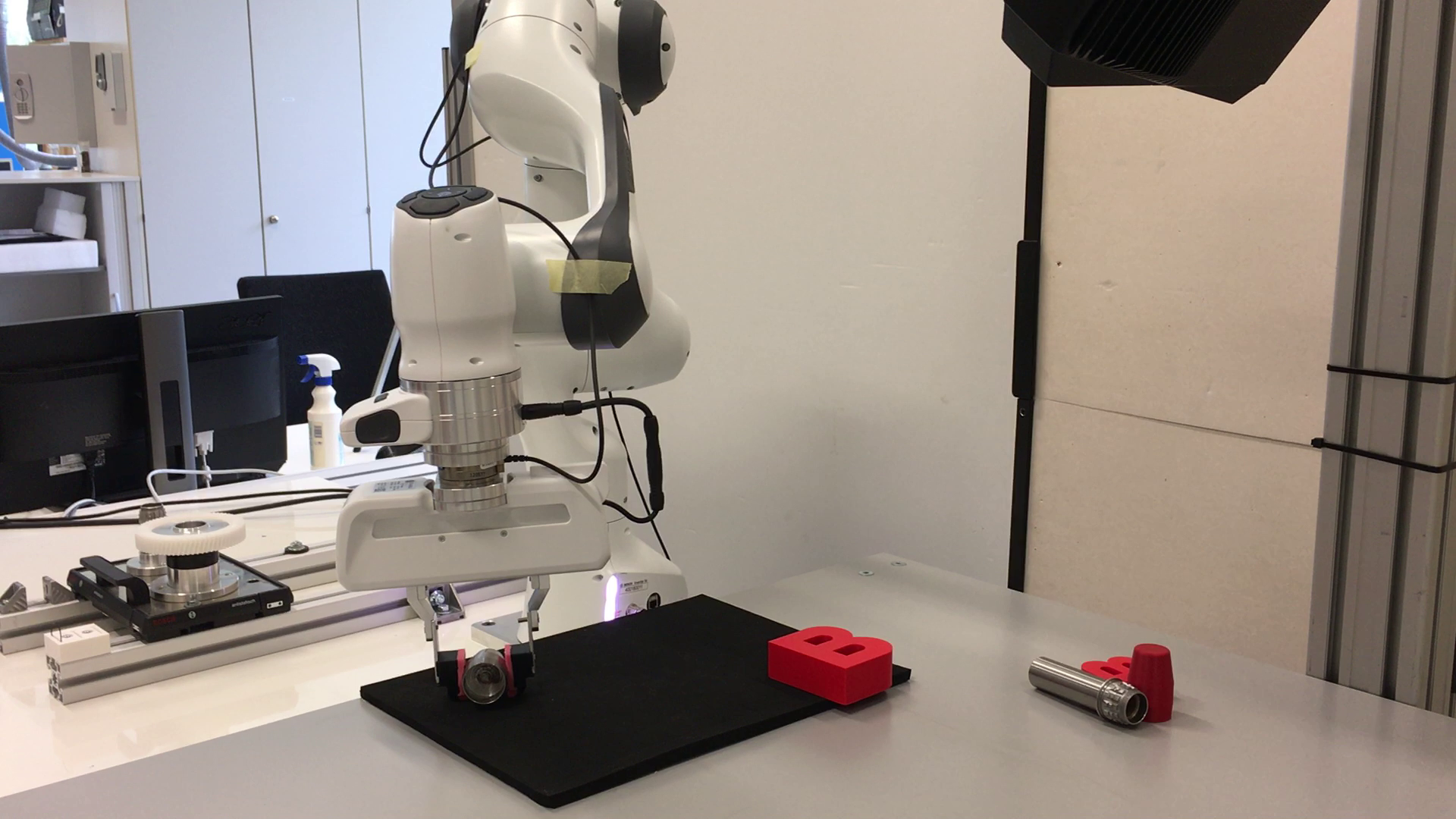}
		\includegraphics[trim={4cm 5cm 32cm 2cm},clip,width=.155\textwidth]{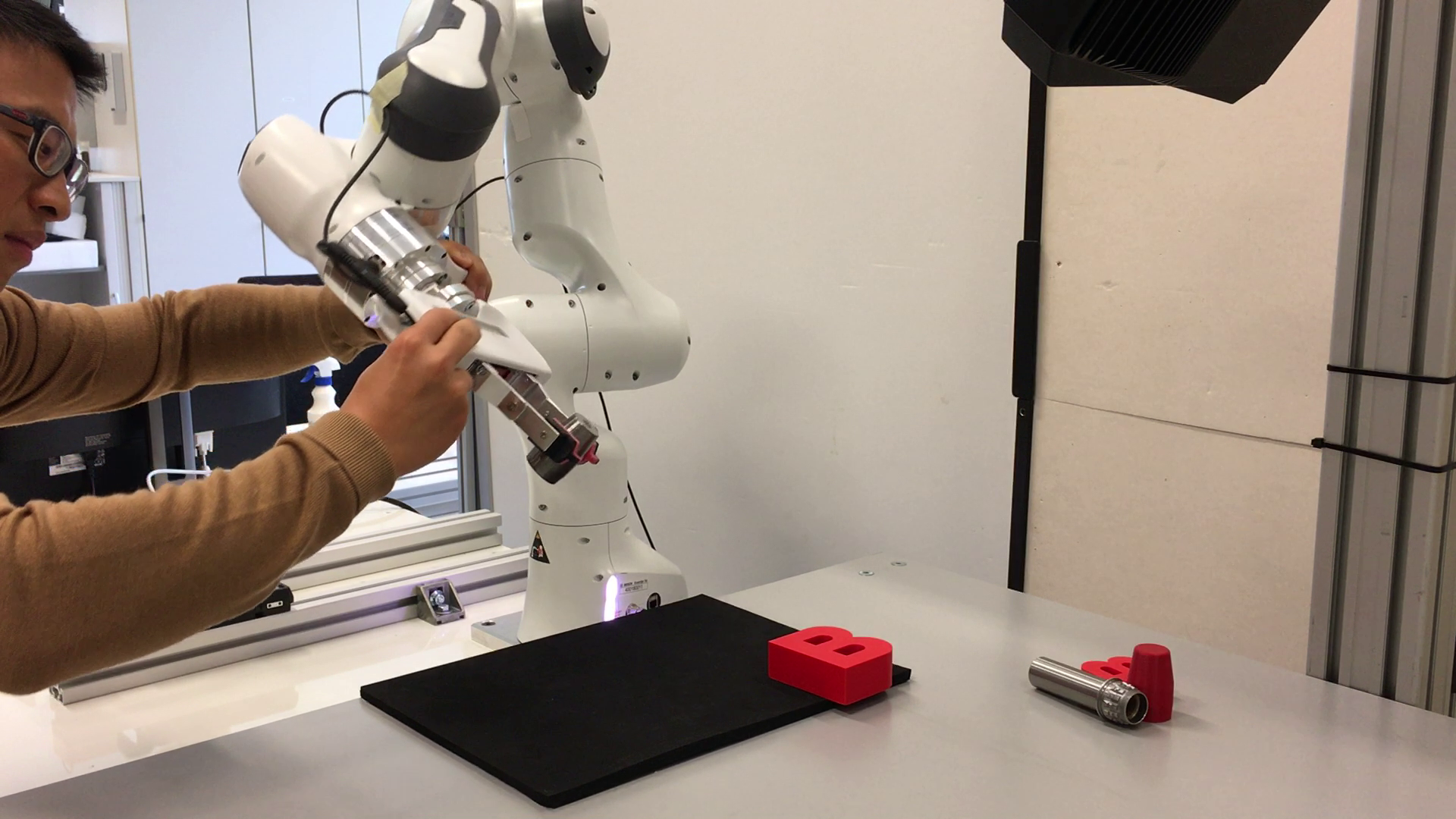}
		\includegraphics[trim={4cm 5cm 32cm 2cm},clip,width=.155\textwidth]{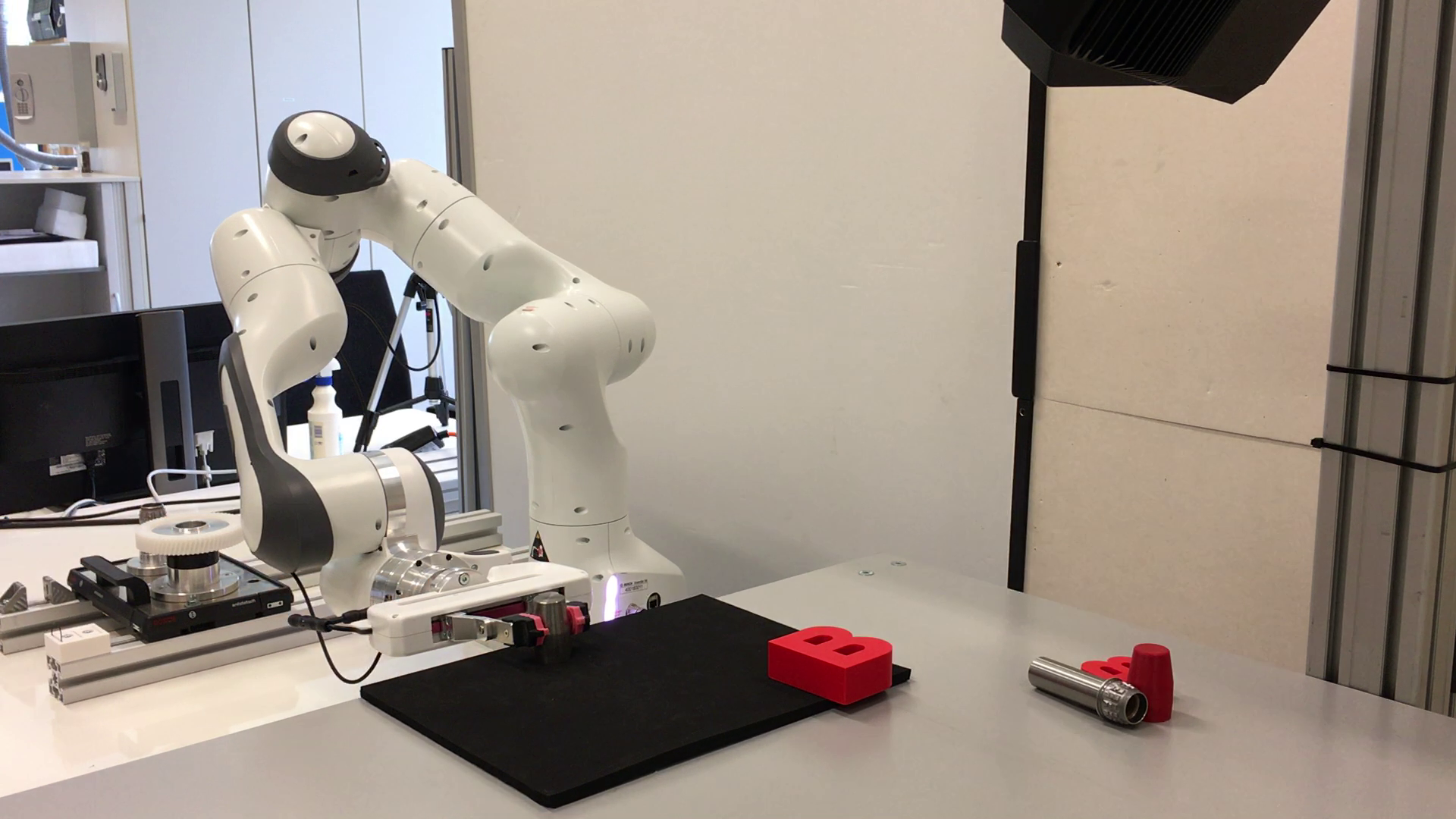}
		\caption{Snapshots of the human demonstrations of the \texttt{re-orient} skill~\cite{Rozo20:Sequencing}}
		\label{Fig:ReorientCapDemos}
		\vspace{-0.35cm}
	\end{figure}

	\paragraph{Manipulation skills on $\mathbb{R}^3 \times \mathcal{S}^3$:}
	To test our approach in a robotic setting, we consider a \texttt{re-orient} skill from~\cite{Rozo20:Sequencing}, which involves lifting a previously-grasped object, rotating the end-effector, and placing the object back on its original location with a modified orientation (see Fig.~\ref{Fig:ReorientCapDemos}). 
	This skill features significant position and orientation changes, and it is therefore suitable to showcase the functionalities of our Riemannian ProMP. 
	We collected $4$ demonstrations of the \texttt{re-orient} skill via kinesthetic teaching on a Franka Emika Panda robot, where full-pose end-effector trajectories $\{\bm{p}_t\}_{t=1}^T$ were recorded, with $\bm{p}_t \in \mathbb{R}^3 \times \mathcal{S}^3$ being the end-effector pose at time step $t$. 
	The data was used to train a ProMP on $\mathbb{R}^3 \times \mathcal{S}^3$, with position and orientation models learned using respectively a classic ProMP and our approach with hyperparameters given in Appendix~\ref{app:experiments}.     
	
\begin{figure*}[t]
		\centering
		\includegraphics[width=\textwidth]{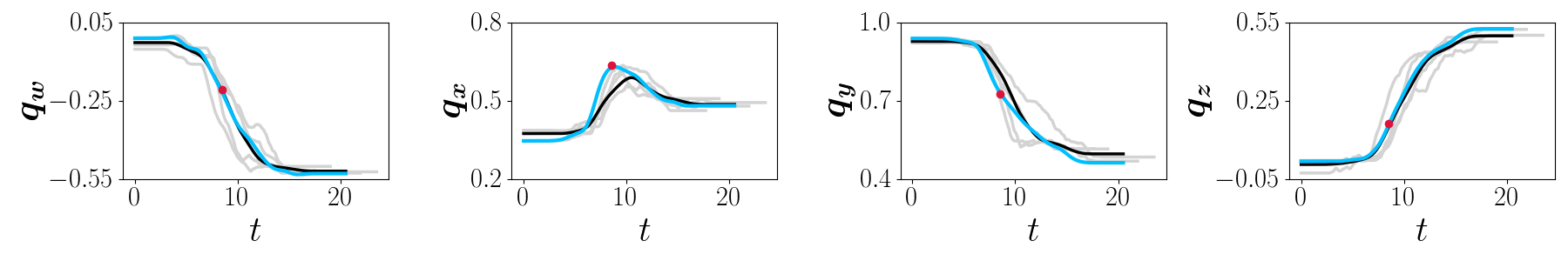}
		\caption{Time-series plot of the \texttt{re-orient} skill demonstrations (\graydemo), original mean trajectory (\blackrepro) of the marginal distribution $\mathcal{P}(\bm{p};\bm{\theta})$, and resulting mean trajectory (\skybluedemo) of the new marginal distribution $\mathcal{P}(\bm{p};\bm{\theta}^*)$ passing through a given via-point (\redcircle) $\bm{p}^* \in \mathbb{R}^3 \times \mathcal{S}^3$. End-effector orientation represented as a quaternion $[\bm{q}_w, \bm{q}_x, \bm{q}_y, \bm{q}_z]$. Time axis given in \si{\sec}.}
		\label{Fig:ReorientCap_ViaPt}
		\vspace{-0.55cm}
	\end{figure*}

	Figure~\ref{Fig:ReorientCap_ViaPt} shows the demonstrations (gray solid lines) and the mean of the marginal distribution $\mathcal{P}(\bm{p};\bm{\theta})$ depicted as a black trajectory. 
	Our model properly captures the motion pattern on $\mathbb{R}^3 \times \mathcal{S}^3$ for this skill. 
	We then evaluated how this learned skill may adapt to a via point $\bm{p}^* \in \mathbb{R}^3 \times \mathcal{S}^3$, representing a new position and orientation of the end-effector at $t=8.5\sec$. 
	By using the approach described in \S~\ref{subsec:condOriProMP}, we computed a new $\mathcal{P}(\bm{p};\bm{\theta}^*)$, where the updated mean is required to pass through $\bm{p}^*$. 
	Figure~\ref{Fig:ReorientCap_ViaPt} displays the updated mean (light blue lines), which successfully adapts to pass through the given via-point. 
	Note that the adapted trajectory exploits the variability of the demonstrations (i.e. the associated covariance) to adapt the trajectory smoothly.
	We also learned two classic ProMPs using an Euler-angle representation and a unit-norm approximation, using the same hyperparameters set. 
	While both models retrieved a distribution $\mathcal{P}(\bm{p};\bm{\theta})$ similar to our approach, their performance is severely compromised in the via-point case, as they retrieve jerky trajectories with lower accuracy tracking w.r.t $\bm{p}^*$ (see Appendix~\ref{subapp:real_exp} and supplemental simulation videos using PyRoboLearn~\cite{Delhaisse19}).
	These results confirm the importance of our Riemannian formulation for ProMP when learning and adapting full-pose end-effector skills. 
	Appendix~\ref{subapp:real_exp} reports learning and reproduction of two additional skills featuring motions of diverse complexity, demonstrating the versatility of our approach.
	
	\section{Discussion}
	\vspace{-0.25cm}
	
	Two main issues were identified in our approach: The effect of the hyperparameters on the trajectory distributions, and the increased computational cost of the weights estimation and blending process. 
	The former problem is a limitation inherited from the classic ProMP, as the more complex the trajectory, the more basis functions we need. 
	The smoothness of this encoding also relies on the basis functions width, which is often the same for all of them. 
	We hypothesize that considering the trajectory curvature to distribute the basis functions accordingly and to define different widths may alleviate this hyperparameter tuning issue.
	The second problem arises as our weights estimation has no closed-form solution, and it relies on Riemannian optimization methods~\cite{Absil07,boumal2020intromanifolds} for solving the geodesic regression and for estimating mean vectors for distributions lying on $\mathcal{M}$. 
	However, note that the weights are estimated only once and offline, meaning that the computations of $\mathcal{P}(\bm{p};\bm{\theta})$ or $\mathcal{P}(\bm{p};\bm{\theta}^*)$ are not compromised. 
	Finally, the gradient-based approach to blend several ProMPs often converges in less than $10$ iterations, making it a fairly fast approach for online blending settings. 
	
	As mentioned in Section~\ref{sec:intro}, several methods overlook the problem of encoding orientation trajectories, or propose approximated solutions that produce distortion and inaccuracy issues (as shown in Appendix~\ref{subapp:real_exp}). 
	Our Riemannian approach explicitly considers the geometric constraints of quaternion data, and therefore it represents a potential alternative to the Riemannian TP-GMM~\cite{Zeestraten17riemannian}. 
	Nevertheless, the benefits of ProMP when compared to TP-GMM lie on the trajectory modulation and blending features, which endow a robot with a larger set of adaptation capabilities. 
	It may be worth investigating how these formal methods dealing with full-pose trajectories compare to each other in terms of accuracy, adaptation, and extrapolation capabilities. 
	In this context, benchmark works similar to~\cite{Rana20benchmarkLfD} may bring important insights to the field.  
	On a different note, the classic ProMP~\cite{paraschos2018probabilistic} also includes time derivatives of the considered variable, which was not covered in this paper. 
	However, this extension is straightforward: This involves to include linear and angular velocities of the end-effector into the trajectory variables. 
	Given that these are Euclidean variables, the main challenge arises when quaternions are part of the trajectory data. 

	
	\clearpage
	\acknowledgments{The authors thank Noémie Jaquier (KIT) and Andras Kupcsik (BCAI) for their support and useful feedback on this work. 
		We would also like to thank the reviewers and area chair for their highly useful recommendations.}
	
	
	\bibliography{References}  
	
	\newpage
	\appendix
	
	\section{Background}
	\label{app:background}
	
	\subsection{Notation}
	\label{sec:notation}
	\begin{table}[h]
		\centering
		\begin{tabularx}{\linewidth}{|l|X|}
			\rowcolor{lightgray}
			\hline
			\textbf{Symbols} & \textbf{Description} \\ [0.3ex] 
			\hline
			$\mathcal{M}$ & Riemannian manifold \\
			$\mathcal{T}_{\bm{\mathrm{p}}}\mathcal{M}$ & Tangent space of manifold $\mathcal{M}$ at $\bm{\mathrm{p}} \in \mathcal{M}$\\
			$\mathcal{T}\mathcal{M}$ & Tangent bundle (group of all the tangent vectors in $\mathcal{M}$)\\
			$\mathcal{N}(\bm{\mu}, \bm{\Sigma})$ & Gaussian distribution with mean $\bm{\mu}\in\mathbb{R}^d$ and covariance~$\bm{\Sigma} ·\in \mathbb{R}^{d\times d}$\\
			$\mathcal{N}_{\mathcal{M}}(\bm{\mu}, \bm{\Sigma})$ & Riemannian Gaussian distribution with mean $\bm{\mu} \in \mathcal{M}$ and covariance $\bm{\Sigma} \in \mathcal{T}_{\bm{\mu}}\mathcal{M}$\\
			$\bm{\theta}$ & Parameters of Gaussian distribution\\
			$\text{Exp}_{\bm{\mathrm{x}}}(\cdot)$ & Exponential map at $\bm{\mathrm{p}}\in \mathcal{M} $ \\
			$\text{Log}_{\bm{\mathrm{x}}}(\cdot)$ & Logarithmic map at $\bm{\mathrm{p}}\in \mathcal{M} $ \\
			$\Gamma_{\bm{\mathrm{x}} \rightarrow \bm{\mathrm{y}}}(\cdot)$ & Parallel Transport from $\mathcal{T}_{\bm{\mathrm{p}}}\mathcal{M}$ to $\mathcal{T}_{\bm{\mathrm{q}}}\mathcal{M}$ \\
			$\mathcal{S}^m$ & $m$-dimensional sphere manifold \\
			$\phi_i(z_t)$ & Normalized Gaussian basis function at time phase $z_t$\\ 
			$\bm{w}$ & ProMP weight vector \\
			$\bm{\Psi}_t$ & ProMP basis function matrix at $z_t$ \\
			\hline
		\end{tabularx}
		\caption{Notation for ProMP and Riemannian manifolds.}
		\label{tab:notation}
		\vspace{-.35cm}	
	\end{table}
	
	\subsection{ProMPs algorithm}
	\label{subapp:promps}
	\begin{algorithm}[h]
		\caption{Classic ProMP Learning}
		\footnotesize
		\SetKwInOut{KwIn}{Input}
		\SetKwInOut{KwIn}{Input}
		\SetKwInOut{KwOut}{Output}
		
		\KwData{ Set of \emph{N} demonstrations with observations 
			$\bm{Y}_n \quad \forall n=1,\ldots,N$.}
		\KwIn{Number of basis functions $N_{\phi}$, width $h$, and regularization term $\lambda$.}
		\KwOut{Mean $\bm{\mu}_{\bm{w}}$ and covariance $\bm{\Sigma_w}$ of $\mathcal{P}(\bm{w};\bm{\theta})$.}
		\ForEach {$demonstration \: n$}
		{
			Compute phase variables: $z_{n} = \frac{t_n}{t_n^{\tiny{final}}}$.\\
			Compute basis functions matrix $\bm{\Psi}_t$ using $\bm{\phi}_m(z_t)$ to build the basis matrix $\bm{\Psi}$.\\
			Compute weight vector $\bm{w}_n$ via $\bm{w}_n = (\bm{\Psi}^{\trsp} \bm{\Psi} +\lambda \mathbf{I})^{-1}\bm{\Psi}^{\trsp} \bm{Y}_n$.
		}
		Fit a Gaussian over the weight vectors $\bm{w}_n$: \vspace*{-0.2cm}
		\begin{align*}
			\bm{\mu_{\bm{w}}} = \frac{1}{N} \sum_{n=1}^{N}{\bm{w}_n},\
			\bm{\Sigma}_{\bm{w}} = \frac{1}{N} \sum_{n=1}^{N}{(\bm{w}_n-\bm{\mu}_{\bm{w}})(\bm{w}_n-\bm{\mu}_{\bm{w}})^\trsp}
		\end{align*}
		\Return $\bm{\mu}_{\bm{w}},\bm{\Sigma}_{\bm{w}}$ \\
		\label{alg:alg1}	
	\end{algorithm}
	
	When learning from demonstrations, the example trajectories often differ in time length. ProMP overcomes this issue by introducing a phase variable $z$ to decouple the data from the time instances, which in turn allows for temporal modulation. In this case, the demonstration ranges from $z_0 = 0$ to $z_T = 1$, redefining the demonstrated trajectory as $\bm{\tau} = \{\bm{y}_t\}_{t = z_0}^{z_T}$. The basis functions that form $\bm{\Psi}$ are defined as a function of the phase variable $z$.
	Specifically, ProMP uses Gaussian basis functions for stroke-based movements, defined as
	\begin{equation}
		b_m(z_t)~=~ \exp\left( \frac{-(z_t-c_m)^2}{2h} \right) \quad \forall m=1,\ldots,N_{\phi} ,
	\end{equation}
	with width $h$ and center $c_i$, which are often experimentally designed.  
	These Gaussian basis functions are then normalized, leading to $\phi_m(z_t)~=~ \frac{b_m(z_t)}{\sum_{j=1}^{N_{\phi}}{b_j(z_t)}}$. 
	The classic ProMP learning process is summarized in Algorithm \ref{alg:alg1}.
	
	\subsection{Riemannian manifolds}
	\label{subapp:riemannian}
	Table~\ref{tab:SphereOperations} provides the different expressions for the Riemannian distance, exponential and logarithmic maps, and parallel transport operation for the sphere manifold $\mathcal{S}^m$. 
	Note that these expressions slightly differ from those used in~\cite{Ude14ICRA}, which actually correspond to exponential and logarithmic maps defined using Lie Theory~\cite{sola2020micro}.
	The main difference lies on the fact that the Lie-based maps are defined with respect to the identity element of the Lie group (the manifold origin, loosely speaking).
	Therefore, they are coupled with transport operations from and to the origin in order to be applied on the complete manifold.
	Here we use maps that are defined at any point $\bm{x} \in \mathcal{M}$.
	Further details can be found when analyzing how the retraction operation is defined using Lie and Riemannian manifolds theories~\cite{sola2020micro,boumal2020intromanifolds}.

	\setlength{\extrarowheight}{2pt}
	\begin{table}[h]
		\centering
		\begin{tabularx}{\linewidth}{|l|X|}
			\rowcolor{lightgray}
			\hline
			\textbf{Operation} & \textbf{Formula} \\ [0.3ex]
			\hline 
			$d_{\mathcal{M}}(\bm{\mathrm{x}},\bm{\mathrm{y}}) $ & $\arccos(\bm{\mathrm{x}}^\trsp \bm{\mathrm{y}})$ \\ [0.3ex] 
			\hline
			$\text{Exp}_{\bm{\mathrm{x}}}(\bm{u})$ & $\bm{\mathrm{x}}\cos(\|\bm{u}\|) + \bm{\overline{u}}\sin(\|\bm{u}\|)$ with $\bm{\overline{u}}=\frac{\bm{u}}{\|\bm{u}\|}$\\ [0.3ex]
			\hline 
			$\text{Log}_{\bm{\mathrm{x}}}(\bm{\mathrm{y}})$ & $ d_{\mathcal{M}}(\bm{\mathrm{x}},\bm{\mathrm{y}}) \, \frac{\bm{\mathrm{y}} - \bm{\mathrm{x}}^\trsp \bm{\mathrm{y}} \, \bm{\mathrm{x}}}{\|\bm{\mathrm{y}} - \bm{\mathrm{x}}^\trsp \bm{\mathrm{y}} \, \bm{\mathrm{x}}\|}$\\ [0.3ex]
			\hline
			$\Gamma_{\bm{\mathrm{x}} \rightarrow \bm{\mathrm{y}}}(\bm{v})$ & $\Big(-\bm{\mathrm{x}}\sin(\|\bm{u}\|)\bm{\overline{u}}^{\trsp} + \bm{\overline{u}}\cos(\|\bm{u}\|)\bm{\overline{u}}^\trsp 
			+ (\bm{I}- \bm{\overline{u}}\,\bm{\overline{u}}^\trsp)\Big)\bm{v}$ with $\bm{\overline{u}}=\frac{\bm{u}}{\|\bm{u}\|}$ and $\bm{u} = \text{Log}_{\bm{\mathrm{x}}}(\bm{\mathrm{y}})$\\ [0.3ex]
			\hline 
		\end{tabularx}
		\caption{Principal operations on $\mathcal{S}^m$ (see~\citet{Absil07} for details).}
		\label{tab:SphereOperations}
		\vspace{-0.35cm}
	\end{table}
	
	We will also need to parallel transport symmetric positive-definite matrices $\bm{\mathrm{V}}$ (e.g. covariance matrices) as follows: $\Gamma_{\bm{\mathrm{x}} \rightarrow \bm{\mathrm{y}}}(\bm{\mathrm{V}}) = \Gamma_{\bm{\mathrm{x}} \rightarrow \bm{\mathrm{y}}}(\bm{l_v})^\trsp \Gamma_{\bm{\mathrm{x}} \rightarrow \bm{\mathrm{y}}}(\bm{l_v})$, where $\bm{l_v}=\operatorname{vec}(\bm{L_v})$ and $\bm{\mathrm{V}} = \bm{L_v}^\trsp\bm{L_v}$. 

	\section{Orientation ProMPs}
	\subsection{Computation of the marginal distribution}
	\label{app:quaternion_promps_marg}
	The marginal distribution of $\bm{\mathrm{y}}_t$ can be computed as\footnote{We drop time index $t$ for the sake of notation.} 
	\begin{equation} \label{eqapp:MarginalDistributionS3}
		\mathcal{P}(\bm{\mathrm{y}};\bm{\theta}) = \int \mathcal{N}_{\tiny{\mathcal{M}}}(\bm{\mathrm{y}} | \underbrace{\text{Exp}_{\bm{\mathrm{p}}}(\bm{\Psi}\bm{w})}_{\bm{\mu_{\mathrm{y}}}}, \bm{\Sigma}_{\bm{\mathrm{y}}}) \mathcal{N}(\bm{w}|\bm{\mu}_{\bm{w}}, \bm{\Sigma}_{\bm{w}}) d\bm{w} ,
	\end{equation}
	where the marginal distribution depends on two probability distributions that lie on different manifolds. However, the mean $\bm{\mu_{\mathrm{y}}}$ depends on a single fixed point $\bm{\mathrm{p}} \in \mathcal{M}$, and $\bm{\mu}_{\bm{w}} \in \mathcal{T}_{\bm{\mathrm{p}}}\mathcal{M}$. We exploit these two observations to solve the marginal \eqref{eqapp:MarginalDistributionS3} on the tangent space $\mathcal{T}_{\bm{\mathrm{p}}}\mathcal{M}$ as follows
	\begin{align*}
		\mathcal{P}(\text{Log}_{\bm{\mathrm{p}}}(\bm{\mathrm{y}})) &= \int \mathcal{N}(\text{Log}_{\bm{\mathrm{p}}}(\bm{\mathrm{y}}) | \bm{\Psi}\bm{w}, \tilde{\bm{\Sigma}}_{\bm{\mathrm{y}}}) \mathcal{N}(\bm{w}|\bm{\mu}_{\bm{w}}, \bm{\Sigma}_{\bm{w}}) d\bm{w} ,\\
		&= \mathcal{N}(\text{Log}_{\bm{\mathrm{p}}}(\bm{\mathrm{y}}) | \bm{\Psi}\bm{\mu}_{\bm{w}}, \bm{\Psi} \bm{\Sigma}_{\bm{w}} \bm{\Psi}^\trsp + \tilde{\bm{\Sigma}}_{\bm{\mathrm{y}}}) ,
	\end{align*} 
	where $\tilde{\bm{\Sigma}}_{\bm{\mathrm{y}}} = \Gamma_{\bm{\mu}_{\bm{\mathrm{y}}} \rightarrow \bm{\mathrm{p}}}(\bm{\Sigma}_{\bm{\mathrm{y}}})$ is the parallel-transported covariance $\bm{\Sigma}_{\bm{\mathrm{y}}}$ from $\bm{\mu}_{\bm{\mathrm{y}}}$ to $\bm{\mathrm{p}}$. Note that this marginal distribution still lies on the tangent space $\mathcal{T}_{\bm{\mathrm{p}}}\mathcal{M}$, so we need to map it back to $\mathcal{M}$ using the exponential map, which leads to the final marginal 
	\begin{equation}\label{eqapp:FinalMarginalOriProMP}
		\mathcal{P}(\bm{\mathrm{y}};\bm{\theta}) = \mathcal{N}_{\tiny{\mathcal{M}}}(\bm{\mathrm{y}} | \underbrace{\text{Exp}_{\bm{\mathrm{p}}}(\bm{\Psi}\bm{\mu}_{\bm{w}})}_{\hat{\bm{\mu}}_{\bm{\mathrm{y}}}}, \hat{\bm{\Sigma}}_{\bm{\mathrm{y}}}), 
	\end{equation}
	where $\hat{\bm{\Sigma}}_{\bm{\mathrm{y}}} = \Gamma_{\bm{\mathrm{p}} \rightarrow \hat{\bm{\mu}}_{\bm{\mathrm{y}}}}(\bm{\Psi} \bm{\Sigma}_{\bm{w}}\bm{\Psi}^\trsp + \tilde{\bm{\Sigma}}_{\bm{\mathrm{y}}})$ is the parallel transportation of the full covariance matrix of the final marginal.

	\subsection{Gradient-based optimization for MGLM}
	\label{subapp:mglm}
	The geodesic regression problem 
	\begin{equation}\label{eqapp:GeodesicRegModel}
		(\hat{\bm{\mathrm{p}}},\hat{\bm{u}}) = \argmin_{(\bm{\mathrm{p}},\bm{u}) \in \mathcal{TM}} \frac{1}{2} \sum_{i=1}^{T}{d_\mathcal{M}(\hat{\bm{\mathrm{y}}}_i,\bm{\mathrm{y}}_i)^2} 
	\end{equation}
	does not yield an analytical form like the original ProMP. As explained by~\citet{Fletcher2013GeodesicRA}, a solution can be obtained through gradient descent, which requires to compute the derivative of the Riemannian distance function and the derivative of the exponential map. The latter is split into derivatives w.r.t the initial point $\bm{\mathrm{p}}$ and the initial velocity $\bm{u}$. These gradients can be computed in terms of Jacobi fields (i.e., solutions to a second order equation subject to certain initial conditions under a Riemannian curvature tensor~\cite{Fletcher2013GeodesicRA}). 
	
	However, to solve the multilinear geodesic regression 
	\begin{equation} \label{eqapp:MultiGeodesicReg}
		(\hat{\bm{\mathrm{p}}},\hat{\bm{u}}_j) = \argmin_{(\bm{\mathrm{p}},\bm{u}_j) \in \mathcal{TM} \, \forall j} \; \frac{1}{2} \sum_{i=1}^{T}{d_\mathcal{M}(\hat{\bm{\mathrm{y}}}_i,\bm{\mathrm{y}}_i)^2} , \quad \text{with} \;  \hat{\bm{\mathrm{y}}}_i = \text{Exp}_{\bm{\mathrm{p}}}(\bm{U}\bm{x}_i) ,
	\end{equation}
	\citet{MGML2014} compute the corresponding gradients by exploiting the insight that the adjoint operators resemble parallel transport operations. In such a way, we can overcome the hurdle of designing special adjoint operators for the multivariate case, and instead, perform parallel transport operations to approximate the necessary gradients. This multivariate framework serves the purpose of our first objective, namely, compute the weight vector for each demonstration lying on a Riemannian manifold $\mathcal{M}$. The weight estimate is here obtained by leveraging~\eqref{eqapp:MultiGeodesicReg}, leading to
	\begin{equation}\label{eqapp:weightOriProMP}
		(\hat{\bm{\mathrm{p}}},\hat{\bm{w}}_m) = \argmin_{(\bm{\mathrm{p}},\bm{w}_m) \in \mathcal{TM} \, \forall m} \; \underbrace{\frac{1}{2} \sum_{t=1}^{T}{d_\mathcal{M}(\text{Exp}_{\bm{\mathrm{p}}}(\bm{W}\bm{\phi}_t), \bm{\mathrm{y}}_t)^2}}_{E(\bm{\mathrm{p}}, \bm{w}_m)} ,
	\end{equation}
	where $\bm{\phi}_t$ is the vector of basis functions at time $t$ and $\bm{W}$ contains the set of estimated tangent weight vectors $\hat{\bm{w}}_m \in \mathcal{T}_{\hat{\bm{\mathrm{p}}}}\mathcal{M}$ (i.e., $N_\phi$ tangent vectors emerging out from the point $\bm{\mathrm{p}} \in \mathcal{M}$), that is, $\bm{W} = \text{blockdiag} (\hat{\bm{w}}_1^\trsp, \ldots, \hat{\bm{w}}_{N_\phi}^\trsp)$. 
	
	To solve~\eqref{eqapp:weightOriProMP}, we need to compute the gradients of $E(\bm{\mathrm{p}}, \bm{w}_m)$ with respect to $\bm{\mathrm{p}}$ and each $\bm{w}_m$. As stated above, these gradients depend on the so-called adjoint operators, which broadly speaking, bring each error term $\text{Log}_{\hat{\bm{\mathrm{y}}}_t}(\bm{\mathrm{y}}_t)$ from $\mathcal{T}_{\hat{\bm{\mathrm{y}}}_t}\mathcal{M}$ to $\mathcal{T}_{\bm{\mathrm{p}}}\mathcal{M}$, with $\hat{\bm{\mathrm{y}}}_t = \text{Exp}_{\bm{\mathrm{p}}}(\bm{W}\bm{\phi}_t)$. Therefore, these adjoint operators can be approximated as parallel transport operations as proposed in~\cite{MGML2014}. This leads to the following reformulation of the error function of~\eqref{eqapp:weightOriProMP} 
	\begin{equation}\label{eqapp:ParallelTransportError}
		E(\bm{\mathrm{p}},\bm{w}_m) = \frac{1}{2} \sum_{t=1}^{T}\| \Gamma_{\hat{\bm{\mathrm{y}}}_t \rightarrow \bm{\mathrm{p}}}( \text{Log}_{\hat{\bm{\mathrm{y}}}_t}(\bm{\mathrm{y}}_t)) \|^2 .
	\end{equation}
	Then, the approximated gradients of the error function $E(\bm{\mathrm{p}},\bm{w}_m)$ correspond to

	\begin{equation}\label{eqapp:GradientsParallelTransportError}
		\nabla_{\bm{\mathrm{p}}} E(\bm{\mathrm{p}},\bm{w}_m) \approx - \sum_{t=1}^{T} \Gamma_{\hat{\bm{\mathrm{y}}}_t \rightarrow \bm{\mathrm{p}}}( \text{Log}_{\hat{\bm{\mathrm{y}}}_t}(\bm{\mathrm{y}}_t)) , \; \text{and} \;
		\nabla_{\bm{w}_m} E(\bm{\mathrm{p}},\bm{w}_m) \approx - \sum_{t=1}^{T} \bm{\phi}_{t,m} \Gamma_{\hat{\bm{\mathrm{y}}}_t \rightarrow \bm{\mathrm{p}}} (\text{Log}_{\hat{\bm{\mathrm{y}}}_t}(\bm{\mathrm{y}}_t)) .
	\end{equation}
	
	The learning process for orientation ProMPs on quaternion trajectories is given in Algorithm~\ref{alg:alg2}. 
	
	\begin{algorithm}[t]
		\caption{Orientation ProMP Learning}
		\footnotesize
		\SetKwInOut{KwIn}{Input}
		\SetKwInOut{KwIn}{Input}
		\SetKwInOut{KwOut}{Output}
		
		\KwData{Set of \emph{N} demonstrations with quaternion data $\bm{Y}_n = \{\bm{\mathrm{y}}_t\}_{t=1}^T\quad \forall n=1,\ldots,N$, with $\bm{\mathrm{y}}_t \in \mathcal{S}^3.$}
		\KwIn{Number of basis functions $N_{\phi}$, width $h$, learning rate $\eta$, maximum learning rate $\eta_{\max}$.}
		\KwOut{Tangent space origin $\bm{\mathrm{p}}$, mean $\bm{\mu}_w$ and covariance $\bm{\Sigma_w}$ of $\mathcal{P}(\bm{w};\bm{\theta})$.}
		\ForEach {$demonstration \: n$}
		{
			$1.$ Compute phase variables: $z_{n} = \frac{t_n}{t_n^{\tiny{final}}}$.\\
			$2.$ Compute basis functions matrix $\bm{\Psi}_t$ using $\bm{\phi}_m(z_t)$ to build the basis matrix $\bm{\Psi}$ of \eqref{eqapp:FinalMarginalOriProMP}.\\
			
			$3.$ \textbf{Gradient descent for MGLM}:\newline
			Initialize: $\bm{\mathrm{p}}_0 \in \mathcal{M}$ (only for $n=1$) and weight vectors $\bm{w}_{m,0}^{(n)} \quad \forall m = 1,\ldots, N_{\phi}$. \newline
			Compute estimated trajectory points $\hat{\bm{\mathrm{y}}}_t = \text{Exp}_{\bm{\mathrm{p}}}(\bm{W}\bm{\phi}_t)$.\newline
			\While {termination criteria}
			{
				$\hat{\bm{\mathrm{p}}} = \text{Exp}_{\bm{\mathrm{p}}_k}(-\eta \nabla_{\bm{\mathrm{p}}}E)$ using \eqref{eqapp:GradientsParallelTransportError}.\\
				$\hat{\bm{w}}_{m} = \Gamma_{\bm{\mathrm{p}}_{k} \rightarrow \bm{\mathrm{p}}_{k+1}}(\bm{w}_{m,k}-\eta \nabla_{\bm{w}_m}E)$ using \eqref{eqapp:GradientsParallelTransportError}.\\
				\eIf{$E(\hat{\bm{\mathrm{p}}}, \hat{\bm{w}}_{m}) < E(\bm{\mathrm{p}}_{k},\bm{w}_{m,k})$}
				{
					$\bm{\mathrm{p}}_{k+1} \leftarrow \hat{\bm{\mathrm{p}}}$\\
					$\bm{w}_{m,k+1} \leftarrow \hat{\bm{w}}_{m}$\\
					$\eta = \min(2\eta,\eta_{\max})$
				}{
					$\eta = \eta/2$
				}
			}
		}
		Fit a Gaussian over the set of concatenated weight vectors $\{\bm{w}^{(n)}\}_{n=1}^N$ with $\bm{w}^{(n)} = \left[\bm{w}_1^{{(n)}^\trsp} \ldots \bm{w}_{N_\phi}^{{(n)}^\trsp}\right]^\trsp$: 
		\begin{align*}
			\bm{\mu_{\bm{w}}} = \frac{1}{N} \sum_{n=1}^{N}{\bm{w}^{(n)}},\
			\bm{\Sigma}_{\bm{w}} = \frac{1}{N} \sum_{n=1}^{N}{(\bm{w}^{(n)}-\bm{\mu}_w)(\bm{w}^{(n)}-\bm{\mu}_w)^\trsp}
		\end{align*}
		\Return $\bm{\mathrm{p}}, \bm{\mu}_{\bm{w}}, \bm{\Sigma}_{\bm{w}}$
		\label{alg:alg2}
	\end{algorithm}

	\subsection{Blending}
	\label{subapp:blendingOriProMP}
	Classic ProMP blends a set of movement primitives by using a product of Gaussian distributions. When it comes to blend primitives in $\mathcal{M}$, one needs to consider that each trajectory distribution is parametrized by a set of weight vectors that lie on different tangent spaces $\mathcal{T}_{\bm{\mathrm{p}}}\mathcal{M}$. We resort to the Gaussian product formulation on Riemannian manifolds introduced by~\citet{Zeestraten17riemannian}, where the log-likelihood of the product is iteratively maximized using a gradient-based approach as proposed by~\citet{Dubbelman11}. 
	Formally, the log-likelihood of a product of Riemannian Gaussian distributions is given by (factoring out constant terms)
	\begin{equation} \label{eq:loglikelihoodPoG}
		\ell(\bm{\mathrm{y}}) = -\frac{1}{2} \sum_{s=1}^S \text{Log}_{\bm{\mu}_{\bm{\mathrm{y}}, s}}(\bm{\mathrm{y}})^{\trsp} \bm{\Sigma}_{\bm{\mathrm{y}}, s}^{-1} \text{Log}_{\bm{\mu}_{\bm{\mathrm{y}}, s}}(\bm{\mathrm{y}}) ,
	\end{equation}
	where $\bm{\mu}_{\bm{\mathrm{y}}, s}$ and $\bm{\Sigma}_{\bm{\mathrm{y}}, s}$ are the parameters of the marginal distribution $\mathcal{P}_s(\bm{\mathrm{y}};\bm{\theta})$ for the skill $s$. Note that the logarithmic maps in~\eqref{eq:loglikelihoodPoG} act on different tangent spaces $\mathcal{T}_{\bm{\mu}_{\bm{\mathrm{y}}, s}}\mathcal{M}, \, \forall \, s=1 \ldots S$. In order to perform the log-likelihood maximization, we need to switch the base and argument of the maps while ensuring that the original log-likelihood function remains unchanged. To do so, we can leverage the relationship $\text{Log}_{\bm{\mathrm{x}}}(\bm{\mathrm{y}}) = -\text{Log}_{\bm{\mathrm{y}}}(\bm{\mathrm{x}})$ as well as parallel transport operations to overcome this problem, leading to
	\begin{equation}\label{eq:ObjectiveFuncPoG}
		\mathcal{J} = \frac{1}{2} \sum_{s=1}^S \text{Log}_{\bm{\mu}^+}(\bm{\mu}_{\bm{\mathrm{y}}, s})^\trsp \bm{\Lambda}_{\bm{\mathrm{y}}, s} \text{Log}_{\bm{\mu}^+}(\bm{\mu}_{\bm{\mathrm{y}}, s})
	\end{equation}
	where $\bm{\mu}^+$ is mean of the resulting Gaussian (that we are estimating) and $\bm{\Lambda}_{\bm{\mathrm{y}}, s} = \Gamma_{\bm{\mu}_{\bm{\mathrm{y}}, s} \rightarrow \bm{\mu}^+}(\bm{\Sigma}_{\bm{\mathrm{y}}, s}^{-1})$. We can rewrite \eqref{eq:ObjectiveFuncPoG} by defining the vector $\epsilon (\bm{\mu}^+)~=~\left[ \text{Log}_{\bm{\mu}^+}(\bm{\mu}_{\bm{\mathrm{y}}, 1})^\trsp  \; \cdots \; \text{Log}_{\bm{\mu}^+}(\bm{\mu}_{\bm{\mathrm{y}}, S})^\trsp\right]^\trsp$ and the block diagonal matrix $\bm{\Lambda} = \text{blockdiag}(\bm{\Lambda}_{\bm{\mathrm{y}}, 1}, \cdots, \bm{\Lambda}_{\bm{\mathrm{y}}, S})$. This results in $\mathcal{J}$ having the form of the objective function used to compute the empirical mean $\bm{v}$ of a Gaussian distribution on a Riemannian manifold $\mathcal{M}$ (Eq. 2.115 in~\citet{Dubbelman11}),
	\begin{equation}
		\label{eq:EmpiricalMeanObjFunction}
		\mathcal{J}(\bm{v}) = \frac{1}{2} \epsilon(\bm{v})^\trsp \bm{\Lambda} \epsilon(\bm{v}), 	
	\end{equation}
	from which is possible to iteratively compute the mean as 
	\begin{equation}
		\bm{v}_{k+1} \leftarrow \text{Exp}_{\bm{v}_{k}}(\Delta_{\bm{v}}) \; \text{with} \;  \Delta_{\bm{v}} = -(\bm{J}^\trsp\bm{\Lambda}\bm{J})^{-1}\bm{J}^\trsp\bm{\Lambda}\epsilon(\bm{v}) ,
	\end{equation}
	where $\bm{J}$ is the Jacobian of $\epsilon(\bm{v})$ with respect to the basis of the tangent space of $\mathcal{M}$ at $\bm{v}_k$. 
	
	\begin{figure*}[t]
		\centering
		\includegraphics[width=.7\textwidth]{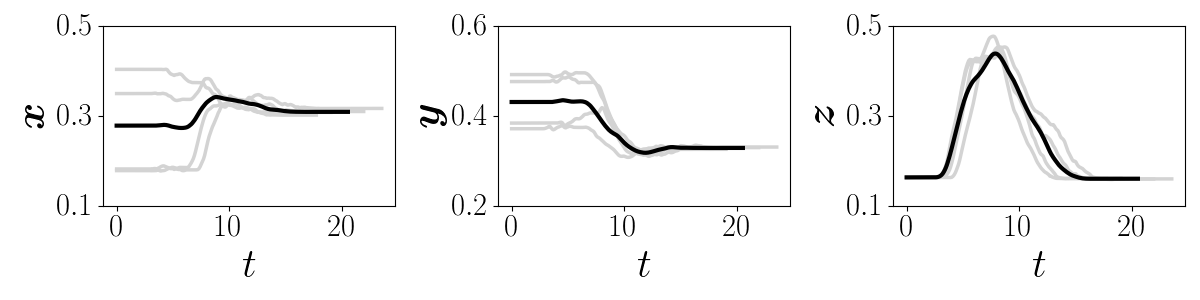}
		\includegraphics[width=\textwidth]{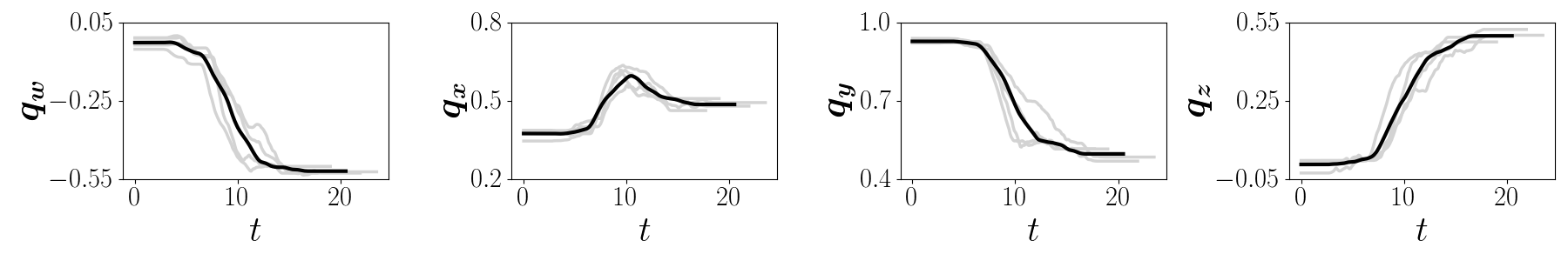}
		\caption{Time-series plot of the \texttt{re-orient} skill demonstrations (\graydemo), and mean trajectory (\blackrepro) of the marginal distribution $\mathcal{P}(\bm{p};\bm{\theta})$. \emph{Top}: end-effector position variables $[x, y, z]$ given in \si{\metre}. \emph{Bottom}: end-effector orientation represented as a quaternion $[\bm{q}_w, \bm{q}_x, \bm{q}_y, \bm{q}_z]$. Time axis given in \si{\sec}.}
		\label{Fig:ReorientCap}
		\vspace{-0.3cm}	
	\end{figure*}

	\begin{figure*}[t]
		\centering
		\includegraphics[width=.7\textwidth]{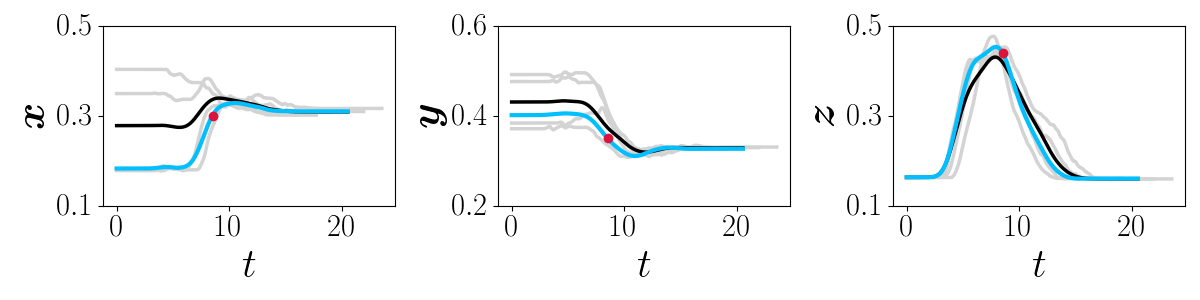}
		\includegraphics[width=\textwidth]{FullPose_viapt_re_orient_cap_right_quat}
		\caption{Time-series plot of the \texttt{re-orient} skill demonstrations (\graydemo), original mean trajectory (\blackrepro) of the marginal distribution $\mathcal{P}(\bm{p};\bm{\theta})$, and resulting mean trajectory (\skybluedemo) of the new marginal distribution $\mathcal{P}(\bm{p};\bm{\theta}^*)$ passing through a given via-point (\redcircle) $\bm{p}^* \in \mathbb{R}^3 \times \mathcal{S}^3$. \emph{Top}: end-effector position variables $[x, y, z]$ given in \si{\metre}. \emph{Bottom}: end-effector orientation represented as a quaternion $[\bm{q}_w, \bm{q}_x, \bm{q}_y, \bm{q}_z]$. Time axis given in \si{\sec}.}
		\label{FigApp:ReorientCap_ViaPt}
		\vspace{-0.3cm}
	\end{figure*}
	
	We can now carry out a similar iterative estimation of the mean $\bm{\mu}^+$ as follows
	\begin{equation}
		\Delta_{\bm{\mu}^+_k} = \left( \sum_{s=1}^{S} \alpha_{s}\bm{\Lambda}_{\bm{\mathrm{y}},s} \right)^{-1} \left(\sum_{s=1}^{S} \alpha_{s}\bm{\Lambda}_{\bm{\mathrm{y}},s} \text{Log}_{\bm{\mu}^+_{k}}(\bm{\mu}_{\bm{\mathrm{y}},s})\right) , \quad \text{and} \quad				
		\bm{\mu}^+_{k+1} \leftarrow \text{Exp}_{\bm{\mu}^+_{k}}(\Delta_{\bm{\mu}^+_k}) , 
	\end{equation}
	where $\bm{\Lambda}_{\bm{\mathrm{y}}, s} = \Gamma_{\bm{\mu}_{\bm{\mathrm{y}}, s} \rightarrow \bm{\mu}^+_k}(\bm{\Sigma}_{\bm{\mathrm{y}}, s}^{-1})$. After convergence at iteration $K$, we obtain the final parameters of the distribution $\mathcal{P}(\bm{y}^+) = \mathcal{N}_{\mathcal{M}}(\bm{y}^+ | \bm{\mu}^{\tiny{+}}, \bm{\Sigma}^{\tiny{+}})$ as follows
	\begin{equation}
		\bm{\mu}^+ \leftarrow \bm{\mu}^+_K \quad \text{and} \quad \bm{\Sigma}^+ = \left(\sum_{s=1}^{S} \alpha_{s}\bm{\Lambda}_{\bm{\mathrm{y}},s} \right)^{-1} .
	\end{equation}
	

	\section{Experiments}
	\label{app:experiments}
	In this section, we provide additional details about the experiments carried out on both the synthetic dataset and the real manipulation skills reported in the main paper. 
	Specifically, we provide the hyperparameters values used to train the models, as well as additional results regarding the comparison against different classical ProMPs implementations: Euler-angles and unit-norm approximations. 
	
	\subsection{Synthetic data experiment on $\mathcal{S}^2$}
	The original trajectories were generated in $\mathbb{R}^2$ and subsequently projected to $\mathcal{S}^2$ by a simple mapping to unit-norm vectors. Each letter in the dataset was demonstrated $N=8$ times. We trained $4$ ProMP models, one for each letter of the set $\{\mathsf{G}, \mathsf{I}, \mathsf{J}, \mathsf{S}\}$. The models trained for $\mathsf{I}$ and $\mathsf{J}$ used $N_{\phi} = 30$ basis functions with width $h=0.01$ and uniformly-distributed centers $c$, while the models trained over the datasets of more involved letters, i.e. $\mathsf{G}$ and $\mathsf{S}$, employed $N_{\phi} = 60$ basis functions with width $h=0.001$. All ProMP models were trained following Algorithm~\ref{alg:alg2} with initial learning rate $\eta = 0.005$, and corresponding upper bound $\eta_{\max} = 0.03$ .           
	
	\begin{figure}[t]
		\centering
		\includegraphics[trim={2cm 3cm 32cm 2cm},clip,width=.17\textwidth]{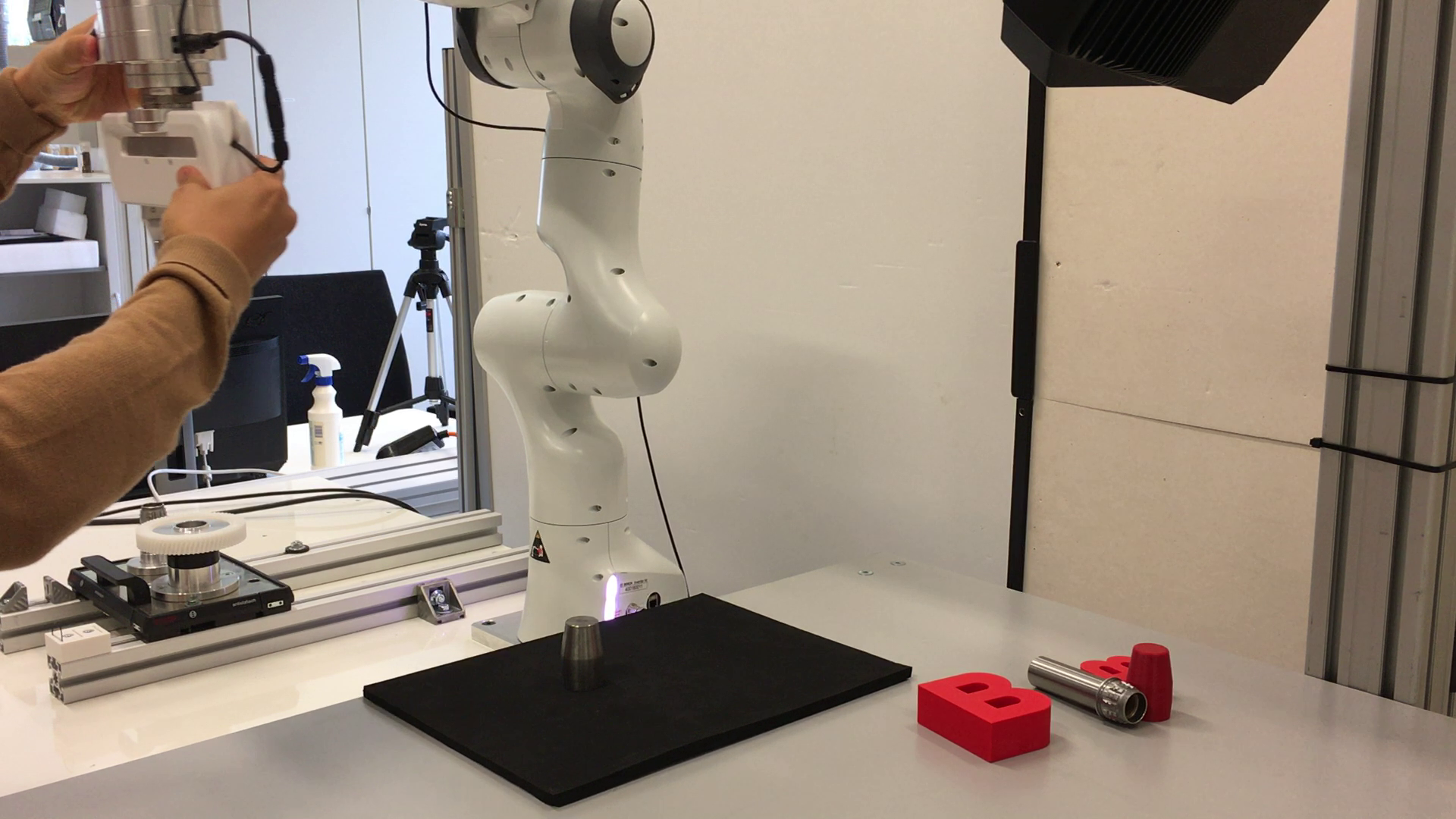}
		\includegraphics[trim={5cm 3cm 29cm 2cm},clip,width=.17\textwidth]{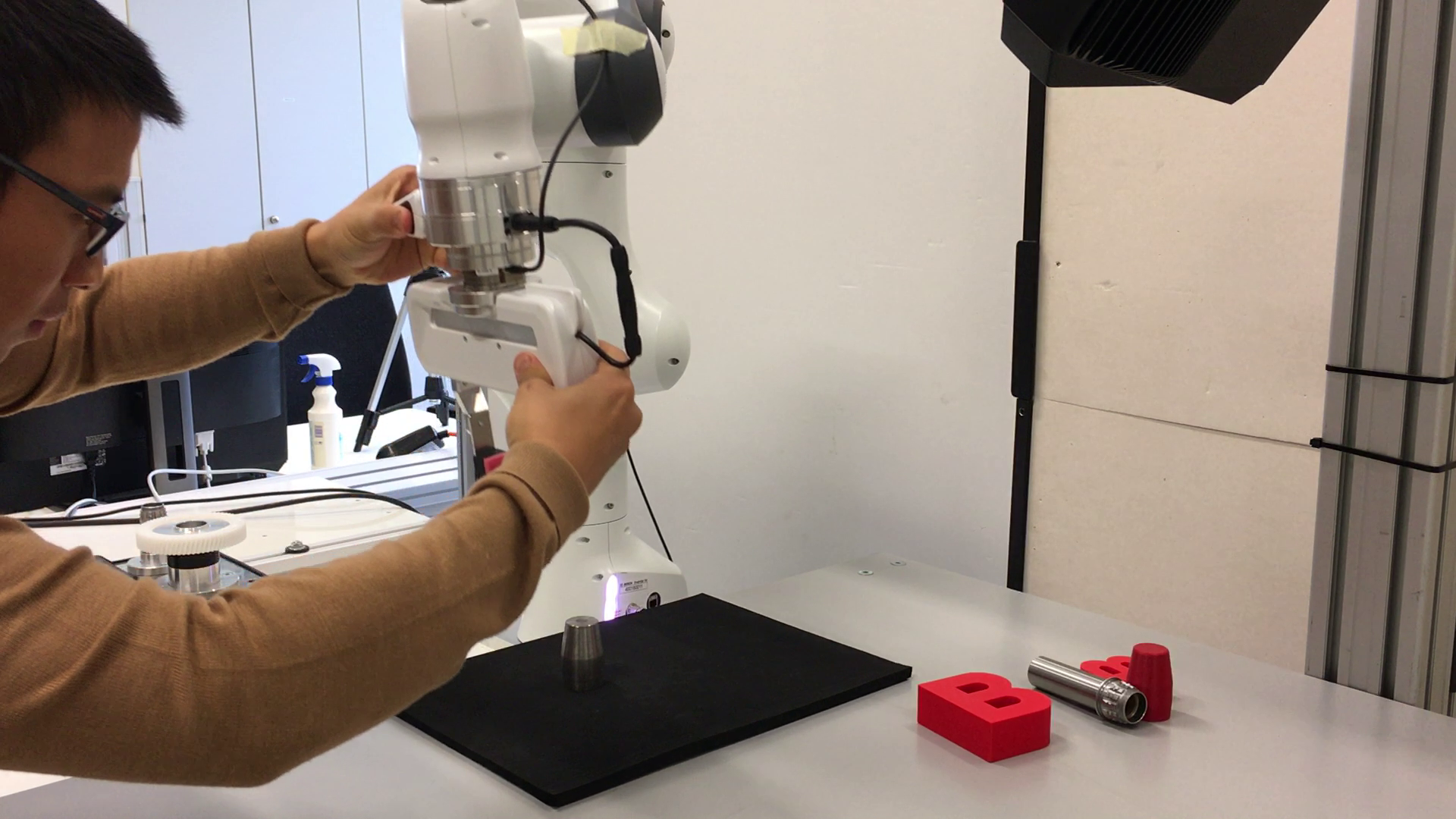}
		\includegraphics[trim={3cm 3cm 29cm 0cm},clip,width=.17\textwidth]{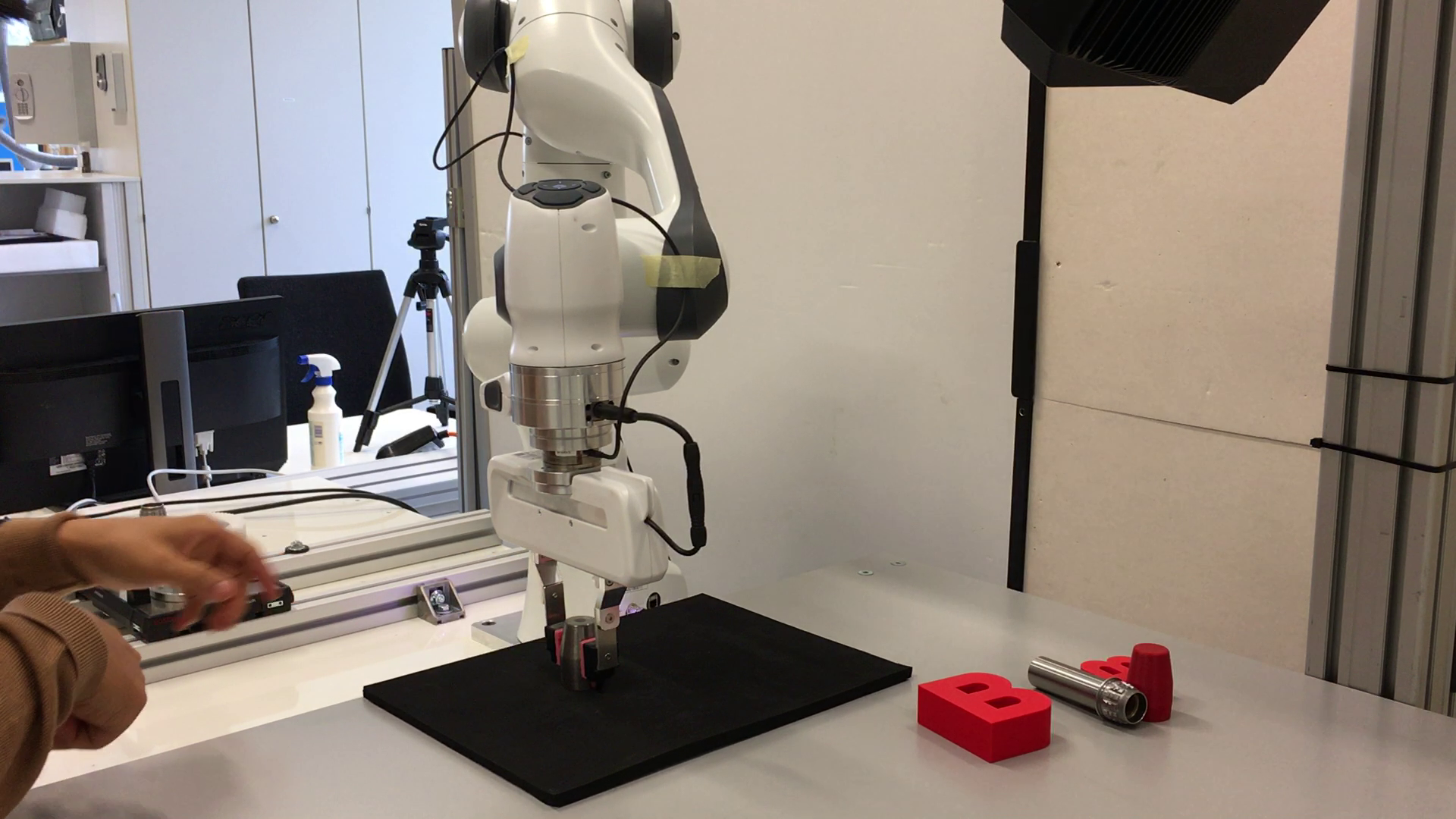}
		\caption{Snapshots of the human demonstrations of the \texttt{top-grasp} skill~\cite{Rozo20:Sequencing}.}
		\label{Fig:TopGraspDemos}
		\vspace{-0.35cm}
	\end{figure}
	
	\begin{figure*}[t]
		\centering
		\includegraphics[width=.7\textwidth]{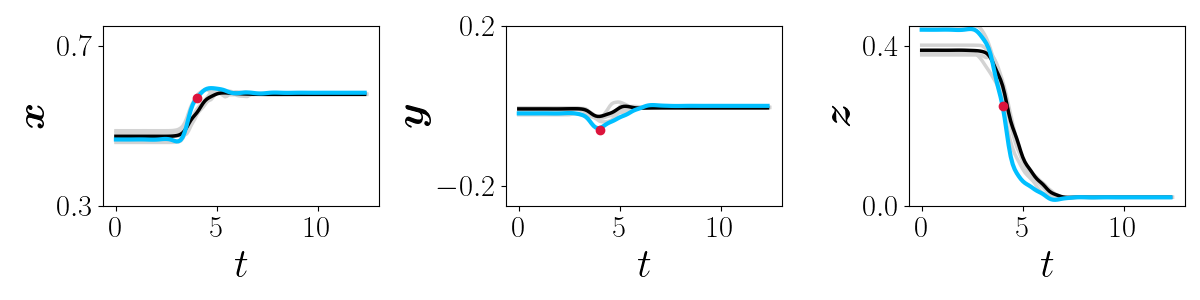}
		\includegraphics[width=\textwidth]{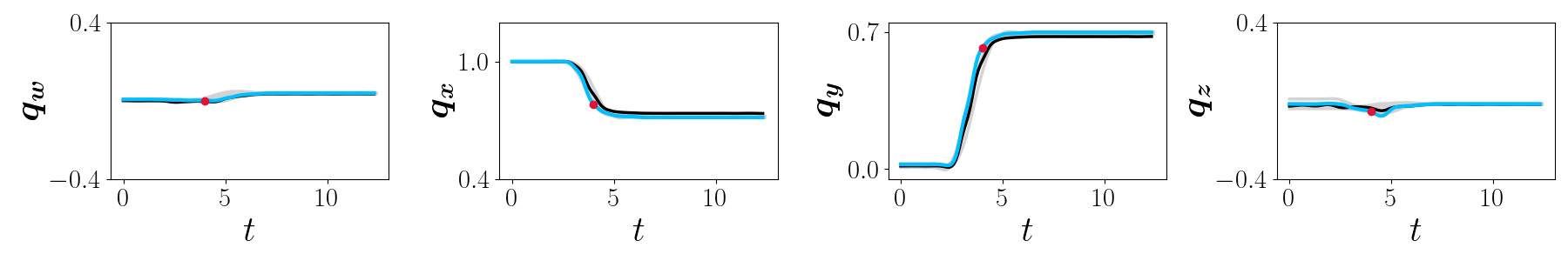}
		\caption{Time-series plot of the \texttt{top-grasp} skill demonstrations (\graydemo), original mean trajectory (\blackrepro) of the marginal distribution $\mathcal{P}(\bm{p};\bm{\theta})$, and resulting mean trajectory (\skybluedemo) of the new marginal distribution $\mathcal{P}(\bm{p};\bm{\theta}^*)$ passing through a given via-point (\redcircle) $\bm{p}^* \in \mathbb{R}^3 \times \mathcal{S}^3$. \emph{Top}: end-effector position variables $[x, y, z]$ given in \si{\metre}. \emph{Bottom}: end-effector orientation represented as a quaternion $[\bm{q}_w, \bm{q}_x, \bm{q}_y, \bm{q}_z]$. Time axis given in \si{\sec}.}
		\label{Fig:TopGrasp_ViaPt}
	\end{figure*}
	
	\subsection{Manipulation skills on $\mathbb{R}^3 \times \mathcal{S}^3$}
	\label{subapp:real_exp}
	\paragraph{Riemannian ProMPs.}
	We collected a set of $4$ demonstrations of the \texttt{re-orient} skill~\cite{Rozo20:Sequencing} through kinesthetic teaching, where full-pose robot end-effector trajectories $\{\bm{p}_t\}_{t=1}^T$ were recorded. Here $\bm{p}_t \in \mathbb{R}^3 \times \mathcal{S}^3$ represents the end-effector pose at time step $t$. The raw data was used to train a Riemannian ProMP on $\mathbb{R}^3 \times \mathcal{S}^3$, where the position and orientation models were learned using Algorithms~\ref{alg:alg1} and~\ref{alg:alg2}, respectively. 
	Both models used the same set of basis functions with $N_{\phi} = 40$, width $h=0.02$, and uniformly-distributed centers $c$. 
	The orientation model was trained using $\eta = 0.005$ as initial learning rate, and corresponding upper bound $\eta_{\max} = 0.03$.     
	
	Figure~\ref{Fig:ReorientCap} shows the recorded demonstrations and the mean of the resulting marginal distribution $\mathcal{P}(\bm{p};\bm{\theta})$. 
	It is evident that the ProMP model properly captures the relevant motion pattern on $\mathbb{R}^3 \times \mathcal{S}^3$. 
	We then evaluated how this learned skill may adapt to a specific via-point. To do so, we chose a via point $\bm{p}^* \in \mathbb{R}^3 \times \mathcal{S}^3$, representing a new position and orientation of the end-effector at $t=8.5\sec$. By using the approach described in Section~\ref{sec:quaternion_promps} of the main paper, we computed a new marginal distribution $\mathcal{P}(\bm{p};\bm{\theta}^*)$, where the updated mean is required to pass through $\bm{p}^*$. Figure~\ref{FigApp:ReorientCap_ViaPt} displays the updated mean of the new marginal distribution, where the original trajectory mean is also displayed for reference. It can be observed that the updated trajectory successfully adapts to pass through the given via-point. Note that the adapted trajectory exploits the variability of the demonstration data (i.e. the associated covariance) to adapt the trajectory smoothly. This adaptation is more pronounced for the position trajectory (top row in Fig.~\ref{FigApp:ReorientCap_ViaPt}), specially along the $x$ axis.

	\begin{figure*}[t]
		\centering
		\includegraphics[width=.7\textwidth]{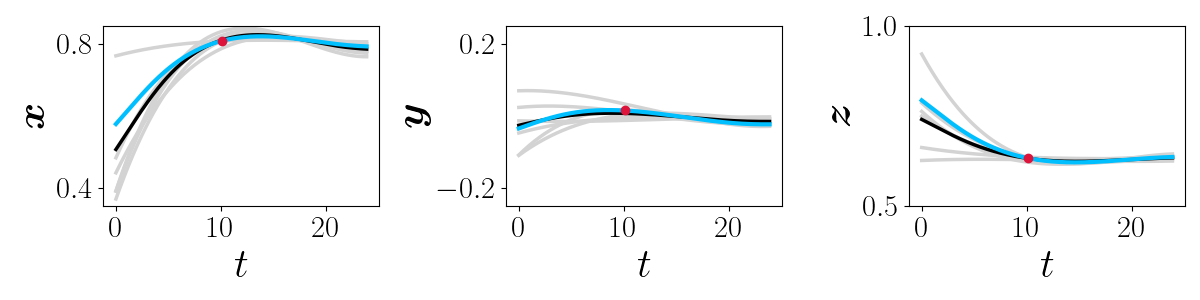}
		\includegraphics[width=\textwidth]{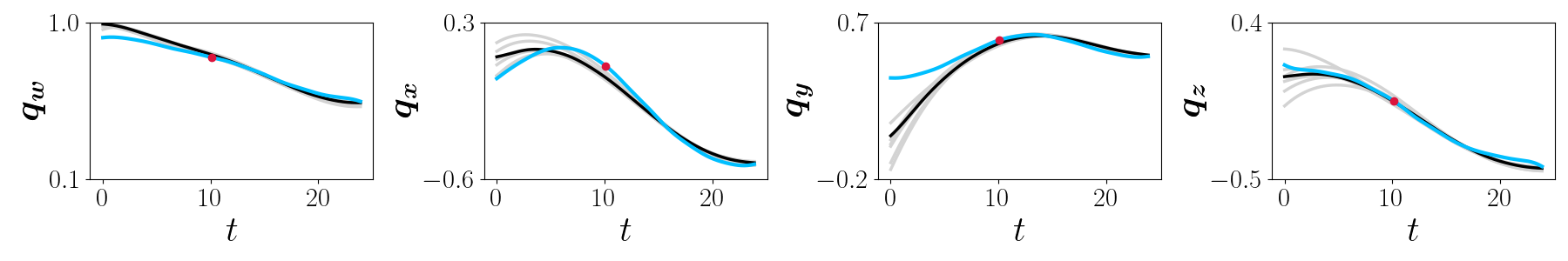}
		\caption{Time-series plot of the \texttt{key-turning} skill demonstrations (\graydemo), original mean trajectory (\blackrepro) of the marginal distribution $\mathcal{P}(\bm{p};\bm{\theta})$, and resulting mean trajectory (\skybluedemo) of the new marginal distribution $\mathcal{P}(\bm{p};\bm{\theta}^*)$ passing through a given via-point (\redcircle) $\bm{p}^* \in \mathbb{R}^3 \times \mathcal{S}^3$. \emph{Top}: end-effector position variables $[x, y, z]$ given in \si{\metre}. \emph{Bottom}: end-effector orientation represented as a quaternion $[\bm{q}_w, \bm{q}_x, \bm{q}_y, \bm{q}_z]$. Time axis given in \si{\sec}.}
		\label{Fig:Synthetic_ViaPt}
		\vspace{-0.3cm}
	\end{figure*}
	
	\begin{figure*}[t]
		\centering
		\textbf{Riemannian ProMP}\par\medskip
		\includegraphics[width=\textwidth]{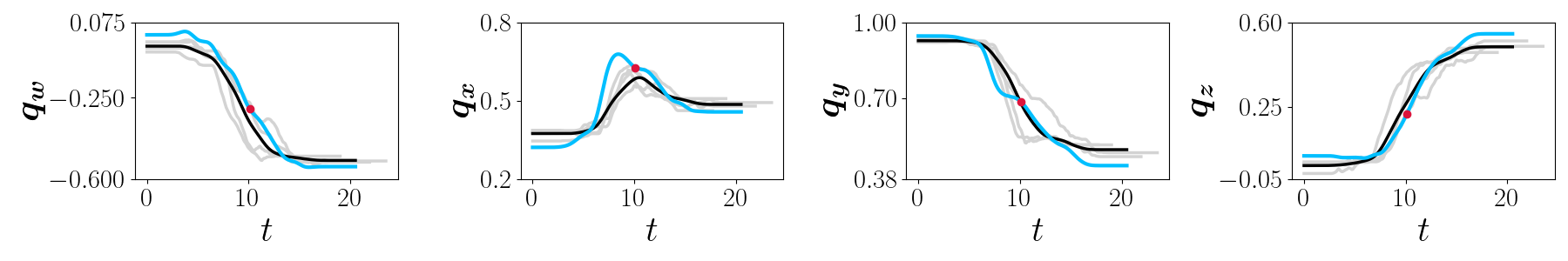}
		\textbf{Euler-angles to quaternion ProMP}\par\medskip
		\includegraphics[width=\textwidth]{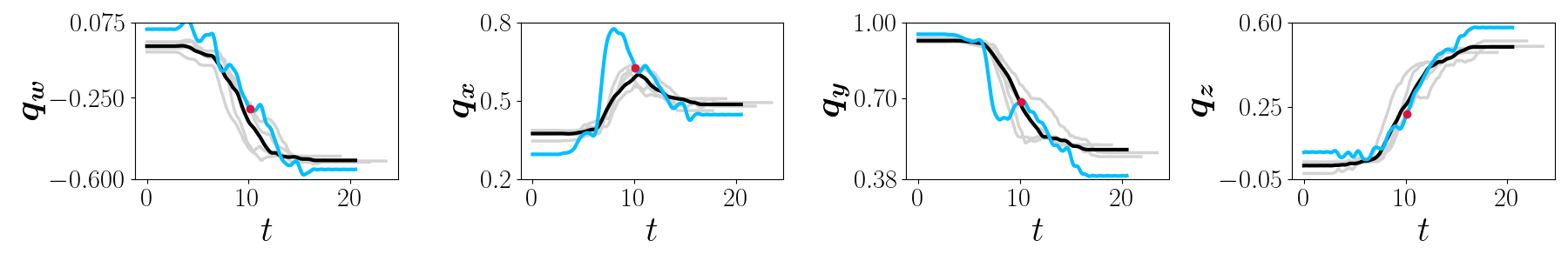}
		\textbf{Unit-norm ProMP}\par\medskip
		\includegraphics[width=\textwidth]{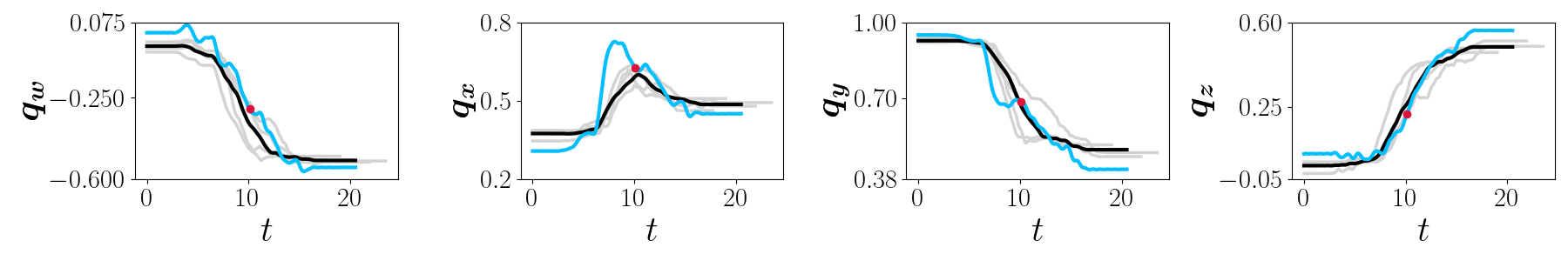}
		\caption{Comparison of mean trajectory retrieval and via-point adaptation for our approach (\emph{top}), Euler-angles ProMPs (\emph{middle}), and unit-norm approximation ProMPs (\emph{bottom}). The figure shows time-series plots of the \texttt{re-orient} skill demonstrations (\graydemo), original mean trajectory (\blackrepro) of the marginal $\mathcal{P}(\bm{p};\bm{\theta})$, and resulting mean trajectory (\skybluedemo) of the new marginal $\mathcal{P}(\bm{p};\bm{\theta}^*)$ passing through the via-point (\redcircle) $\bm{p}^* \in \times \mathcal{S}^3$. The end-effector orientation is represented as a quaternion $[\bm{q}_w, \bm{q}_x, \bm{q}_y, \bm{q}_z]$. Time axis given in \si{\sec}.}
		\label{Fig:ComparedApproaches}
	\end{figure*}

	To showcase the versatility of our approach, we trained two additional Riemannian ProMPs over data collected for the \texttt{top-grasp} and \texttt{key-turning} skills. 
	The former corresponds to a dataset recorded on the same robotic setup as that of the \texttt{re-orient} skill~\cite{Rozo20:Sequencing}, while the latter is a synthetic dataset recorded from a robotic simulator.
	Figure~\ref{Fig:TopGraspDemos} displays the demonstration process of the \texttt{top-grasp} skill: Approach the metallic cap and grasp it from its top. 
	Four demonstrations were used to train a Riemannian ProMP using a set of basis functions with $N_{\phi} = 20$, width $h=0.025$, and uniformly-distributed centers $c$. 
	Figure~\ref{Fig:TopGrasp_ViaPt} shows our model successfully learns the relevant motion pattern of this skill.
	We also tested our approach on a dataset of synthetic demonstrations simulating a \texttt{key-turning} skill: Approach the key hole while aligning the end-effector accordingly, and later turning the end-effector $90$ degrees clock-wise.
	Figure~\ref{Fig:Synthetic_ViaPt} displays the recorded demonstrations, the mean of the resulting marginal distribution, and updated mean trajectory for a given via-point. 
	Our model once again captures the motion patterns for the \texttt{key-turning} skill, and exploits the demonstrations variability to update the mean trajectory while adapting to a via-point.
	These results confirm that our Riemannian formulation for ProMP makes full-pose trajectory learning and adaptation possible. 
	The reproduction of the learned skills in simulation, using PyRoboLearn~\cite{Delhaisse19}, can be watched in the supplementary video at \href{https://sites.google.com/view/orientation-promp}{https://sites.google.com/view/orientation-promp}.
	
	\setlength{\extrarowheight}{2pt}
	\begin{table}[t]
		\centering
		\begin{tabularx}{\linewidth}{|l|X|X|X|}
			\rowcolor{lightgray}
			\hline
			& \textbf{Riemannian ProMPs} & \textbf{Euler ProMPs} & \textbf{Unit-norm ProMPs}\\ [0.3ex]
			\hline 
			Jerkiness & $\bm{21.1}$ & $338.8$ & $278.7$ \\ [0.3ex] 
			\hline
			Tracking accuracy & $\bm{6.25\times10^{-5}}$ & $1.28\times10^{-3}$ & $6.24\times10^{-4}$ \\ [0.3ex]
			\hline 
			Deviation from mean & $\bm{18.16}$ & $28.15$ & $22.94$ \\ [0.3ex]
			\hline 
		\end{tabularx}
		\caption{Trajectory jerkiness (a.k.a smoothness~\cite{Balasubramanian2015:smoothness}), tracking accuracy, and deviation from the mean trajectory for via-point adaptation of the \texttt{re-orient} skill. Bold values represent the best achieved result.}
		\label{tab:ApproachesPerformance}
	\end{table}

	\setlength{\extrarowheight}{2pt}
	\begin{table}[t]
		\centering
		\begin{tabularx}{\linewidth}{|l|X|X|X|}
			\rowcolor{lightgray}
			\hline
			& \textbf{Riemannian ProMPs} & \textbf{Euler ProMPs} & \textbf{Unit-norm ProMPs}\\ [0.3ex]
			\hline 
			Jerkiness & $\bm{114.85}$ & $956.2$ & $523.1$ \\ [0.3ex] 
			\hline
			Tracking accuracy & $\bm{1.0\times10^{-7}}$ & $9.8\times10^{-4}$ & $2.3\times10^{-4}$ \\ [0.3ex]
			\hline 
			Deviation from mean & $\bm{2.43}$ & $5.2$ & $4.1$ \\ [0.3ex]
			\hline 
		\end{tabularx}
		\caption{Trajectory jerkiness (a.k.a smoothness~\cite{Balasubramanian2015:smoothness}), tracking accuracy, and deviation from the mean trajectory for via-point adaptation of the \texttt{top-grasp} skill. Bold values represent the best achieved result.}
		\label{tab:ApproachesPerformanceTopGrasp}
	\end{table}
	
	\setlength{\extrarowheight}{2pt}
	\begin{table}[t]
		\centering
		\begin{tabularx}{\linewidth}{|l|X|X|X|}
			\rowcolor{lightgray}
			\hline
			& \textbf{Riemannian ProMPs} & \textbf{Euler ProMPs} & \textbf{Unit-norm ProMPs}\\ [0.3ex]
			\hline 
			Jerkiness & $\bm{2.52}$ & $28.8$ & $18.9$ \\ [0.3ex] 
			\hline
			Tracking accuracy & $\bm{7.0\times10^{-4}}$ & $2.8\times10^{-3}$ & $9.8\times10^{-4}$ \\ [0.3ex]
			\hline 
			Deviation from mean & $\bm{19.63}$ & $30.5$ & $24.4$ \\ [0.3ex]
			\hline 
		\end{tabularx}
		\caption{Trajectory jerkiness (a.k.a smoothness~\cite{Balasubramanian2015:smoothness}), tracking accuracy, and deviation from the mean trajectory for via-point adaptation of the \texttt{key-turning} skill. Bold values represent the best achieved result.}
		\label{tab:ApproachesPerformanceKeyTurning}
	\end{table}

	\paragraph{Comparison against Euler-angles and unit-norm ProMPs.}
	Using the same demonstrations of the aforementioned skills, we trained three different ProMPs over the orientation trajectories to assess the importance of considering the quaternion geometry as proposed in this paper. 
	The first model corresponds to our Riemannian ProMP approach, the second model is a classical ProMP trained over the Euler angles of the recorded orientation trajectories, and the third model is a classical ProMP trained over quaternion trajectories, whose output is normalized to comply with the unit-norm constraint. 
	All models used the same set of hyperparameters detailed above. 
	Moreover, our Riemannian models were trained using $\eta = 0.005$ as initial learning rate, and corresponding upper bound $\eta_{\max} = 0.03$.
	The output distributions from the Euler-angles model were transformed to quaternion trajectories mainly for comparison purposes. 
	However, the robot orientation controller also works with quaternion data, therefore such transformation is required whenever we need to send the desired orientation trajectory to the robot.       
	
	\setlength{\extrarowheight}{2pt}
	\begin{table}[t]
		\centering
		\begin{tabularx}{\linewidth}{|l|X|X|X|}
			\rowcolor{lightgray}
			\hline
			& \textbf{Riemannian ProMPs} & \textbf{Euler ProMPs} & \textbf{Unit-norm ProMPs}\\ [0.3ex]
			\hline 
			Weights estimation & $254.8\sec$ & $1.8097\sec$ & $3.2797\sec$ \\ [0.3ex] 
			\hline
			Trajectory retrieval & $0.2266 \sec$ & $0.2089\sec$ & $0.2391\sec$ \\ [0.3ex]
			\hline 
			Via-point conditioning & $0.3261 \sec$ & $0.2321\sec$ & $0.2481\sec$ \\ [0.3ex]
			\hline 
		\end{tabularx}
		\caption{Comparison of the computational cost when learning the model weights and retrieving the trajectory distribution for the standard and via-point cases of the $\texttt{re-orient}$ skill. Trajectory retrieval and via-point conditioning times consider the computation cost over the whole trajectory.}
		\label{tab:ComputationCost}
	\end{table}
	
	Figure~\ref{Fig:ComparedApproaches} shows the mean of the resulting marginal (black line) and via-point adapted (light-blue line) distributions for each approach trained over the \texttt{re-orient} skill.
	No significant differences were observed when comparing the mean trajectories of the marginal distributions $\mathcal{P}(\bm{\mathrm{y}};\bm{\theta})$ and $\mathcal{P}(\bm{y};\bm{\theta})$, i.e. Riemannian and classical ProMPs, respectively.
	However, when we evaluated how these models adapt to via-points (e.g. at $t=10.0\sec$), the importance of considering the quaternion space geometry is very noticeable.
	First, the deviation from the original mean $\sum_{t=1}^T d_{\mathcal{M}}(\bm{\mathrm{y}}_t,\bm{\mathrm{y}}_t^*)$ significantly increases for models trained over Euler angles and unit-norm approximation. 
	Second, the tracking accuracy w.r.t the via-point $d_{\mathcal{M}}(\bm{\mathrm{y}}_{t=10},\bm{\mathrm{y}}_{t=10}^*)$ is compromised when using the Euclidean approaches. 
	Third, our Riemannian approach retrieves the smoothest adapted trajectories when compared to its Euclidean counterparts, which is of paramount important when controlling real robots (supplemental simulation videos using PyRoboLearn~\cite{Delhaisse19} can be found at \href{https://sites.google.com/view/orientation-promp}{https://sites.google.com/view/orientation-promp}).  
	Quantitative measures regarding trajectory smoothness, accuracy and deviation are given in Table~\ref{tab:ApproachesPerformance}, where it is clear that our Riemannian formulation outperforms the other two Euclidean methods. 
	Similar results were obtained for both the \texttt{top-grasp} and \texttt{key-turning} skills, reported in Tables~\ref{tab:ApproachesPerformanceTopGrasp} and~\ref{tab:ApproachesPerformanceKeyTurning}, respectively.
	
	As discussed in the main paper, the aforementioned benefits of our Riemannian approach comes at the cost of increasing the computational complexity of the model weights estimation.
	Table~\ref{tab:ComputationCost} reports the computational cost for our Riemannian formulation, the Euler and Unit-norm ProMPs for the \texttt{re-orient} skill. 
	We measured the execution time for the model weights estimation, trajectory retrieval and via-point conditioning. 
	It is evident that weights estimation based on multivariate geodesic regression takes significantly longer than the original ProMP approach. 
	Note that this execution time can be easily reduced by relaxing the stopping criteria of the gradient-based optimizer, such as maximum iterations and geodesic reconstruction error. 
	More importantly, the trajectory retrieval and via-point conditioning processes are not compromised, which are the methods that a practitioner may be interest in running on the real platform. 
	All the approaches were implemented in unoptimized Python code.

\end{document}